\documentclass{article} 
\usepackage{iclr2027conference,times}

\usepackage{amsmath,amsfonts,bm}

\def\eqref#1{equation~\ref{#1}}

\def\1{\bm{1}}

\DeclareMathAlphabet{\mathsfit}{\encodingdefault}{\sfdefault}{m}{sl}
\SetMathAlphabet{\mathsfit}{bold}{\encodingdefault}{\sfdefault}{bx}{n}

\usepackage{graphicx}
\usepackage{hyperref}
\usepackage{url}
\usepackage{hyperref}
\usepackage[normalem]{ulem}
\usepackage{dsfont}
\usepackage{url}
\usepackage{graphicx}
\usepackage{xspace}
\usepackage{makecell}
\usepackage{array}
\usepackage{colortbl}
\usepackage{multirow}
\usepackage{subcaption}
\usepackage{tabularx}
\usepackage{color}
\usepackage{pifont}
\usepackage[utf8]{inputenc}
\usepackage[T1]{fontenc}
\usepackage{CJKutf8}
\usepackage{enumitem}
\usepackage{diagbox}
\usepackage{listings}
\usepackage[flushleft]{threeparttable}
\usepackage{booktabs}
\usepackage{wrapfig}
\usepackage{algorithmicx,algorithm}
\usepackage{graphicx}
\usepackage{enumitem}
\usepackage{caption}
\usepackage{placeins}
\usepackage{needspace}
\usepackage{booktabs}
\usepackage{array}
\usepackage{adjustbox}
\usepackage{xcolor}

\usepackage[most]{tcolorbox}
\usepackage{xcolor}
\usepackage{graphicx}
\usepackage{subcaption}
\usepackage{amsmath,amssymb}
\usepackage{iftex}
\usepackage{marvosym}

\usepackage{iftex}

\ifPDFTeX
    \usepackage[T1]{fontenc}
    \usepackage[type1,nosfdefault]{comicneue}

    \newcommand{\generationfont}{\comicneue}

\else
    \usepackage{fontspec}

    \IfFontExistsTF{Comic Sans MS}
        {\newfontfamily\generationfont{Comic Sans MS}}
        {\newfontfamily\generationfont{TeX Gyre Adventor}}
\fi

\definecolor{GenBlue}{HTML}{315BAA}
\definecolor{GenTextBlue}{HTML}{4677CF}
\definecolor{GenOrange}{HTML}{F07C22}
\definecolor{GenGreen}{HTML}{2A9D64}
\definecolor{GenGray}{HTML}{4E5969}

\newtcolorbox{generationcase}[1][]{
  enhanced,
  colback=white,
  colframe=GenBlue,
  boxrule=1.15pt,
  arc=4mm,
  left=3mm,right=3mm,top=2.5mm,bottom=2.5mm,
  fontupper=\generationfont\small,
  #1
}
\newcommand{\templatefield}[1]{%
  \textcolor{GenOrange}{\textbf{Template:}} #1\par\smallskip}
\newcommand{\seriesfield}[1]{%
  \textcolor{GenGreen}{\textbf{Example Series:}} #1\par\smallskip}
\newcommand{\generatedfield}[1]{%
  \textcolor{GenTextBlue}{\textbf{Generated Text:}} #1}

\newcommand{\cmark}{\textcolor{green!60!black}{\ding{51}}}
\newcommand{\xmark}{\textcolor{red!80!black}{\ding{55}}}

\newcommand{\name}{{QiYao-M}}

\definecolor{fbApp}{HTML}{ffe4e3}
\newcommand{\rowcg}{\rowcolor{gray!10}}

\newcommand{\methodlogo}[2]{%
    #2\hspace{0.15em}%
    \raisebox{-0.25\height}{\includegraphics[height=1.25em]{#1}}%
}

\definecolor{bestpurple}{RGB}{230,220,255}
\definecolor{secondblue}{RGB}{220,235,255}
\newcommand{\best}[1]{\cellcolor{bestpurple}\textbf{#1}}
\newcommand{\second}[1]{\cellcolor{secondblue}\underline{#1}}

\newcommand{\rkw}[1]{%
  \begingroup
  \setlength{\fboxsep}{1pt}%
  \colorbox{bestpurple}{{\textbf{#1}}}%
  \endgroup
}

\newcommand{\rjw}[1]{%
  \begingroup
  \setlength{\fboxsep}{1pt}%
  \colorbox{secondblue}{{\underline{#1}}}%
  \endgroup
}

\title{\name{}: Multimodal Time Series Foundation Model with Role-Aware Modeling of Endogenous and Exogenous Modalities}

\author{
Hanyin Cheng$^{1,*}$,
Linfeng Wang$^{1,*}$,
Zhengbo Qu$^{1}$,
Yang Shu$^{1}$,
Zhongwen Rao$^{2}$\textsuperscript{\Letter}, \\
~\textbf{Meng Wang}$^{2}$\textbf{,} 
\textbf{Yijie Li}$^{2}$\textbf{,}
\textbf{Xin Jiang}$^{2}$\textbf{,}
\textbf{Bin Yang}$^{1}$\textbf{,}
\textbf{Chenjuan Guo}$^{1}$\textsuperscript{\Letter} \\
$^{1}$East China Normal University \\
$^{2}$Huawei Technologies Co., Ltd. \\
\texttt{\{hycheng,lfwang,zbqu07\}@stu.ecnu.edu.cn,} \\
\texttt{\{raozhongwen,wangmeng71,liyijie5,Jiang.Xin\}@huawei.com,} \\
\texttt{\{yshu,byang,cjguo\}@dase.ecnu.edu.cn}
}

\iclrfinalcopy 
\begin{document}

\maketitle
\begingroup
\renewcommand{\thefootnote}{}
\footnotetext{%
$^{*}$ Equal contribution. \quad
\Letter~Corresponding authors.
}
\addtocounter{footnote}{-1}
\endgroup
\lhead{Preprint}
\begin{abstract}

Existing multimodal time series foundation models (TSFMs) typically model
heterogeneous modalities through largely shared mechanisms, overlooking the
distinct forecasting roles of endogenous and exogenous modalities.
In this work, we propose \textbf{QiYao-M}, a role-aware multimodal TSFM that models the two types of modalities separately. 
For endogenous modalities, to capture how they evolve along with the underlying temporal dynamics, we introduce an \textit{Endo-Multimodal Predictor} and \textit{Endo-Multimodal Supervision} to explicitly learn their evolution from history to the future.
For exogenous modalities, to generalize across domains and across various modality types and numbers under the scarcity of exo-multimodal pretraining data, we propose an \textit{Exo-Multimodal Retrieval Enhancer} that enables rapid downstream adaptation without updating the TSFM parameters. We further introduce \textit{Endo-Modality Proxy Training} to train this retrieval module without exogenous multimodal pretraining data.  Extensive experiments across unimodal and multimodal benchmarks demonstrate strong forecasting performance in scenarios both with and without exogenous modalities.

\end{abstract}

\section{Introduction}

\begin{figure}[!thbp]
\includegraphics[width=1\linewidth]{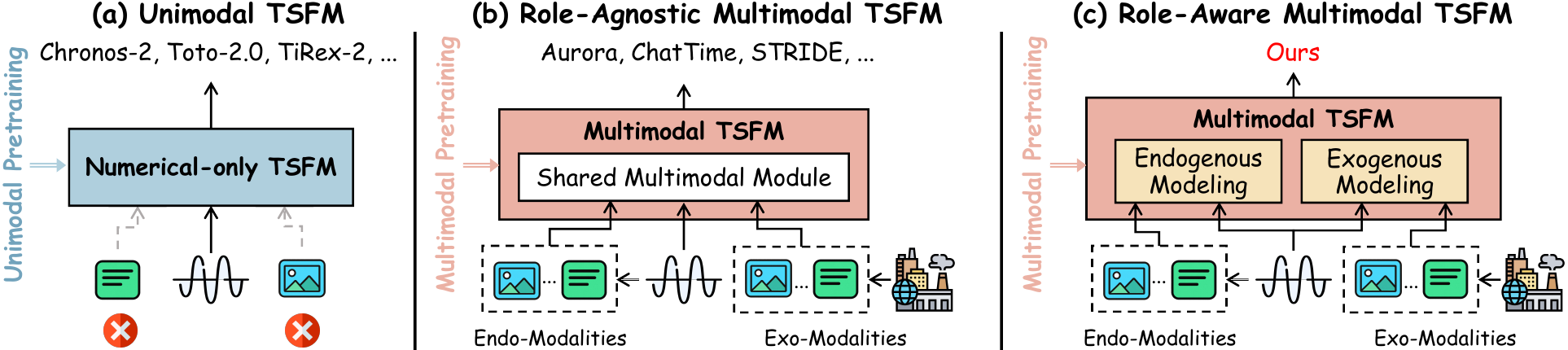}
  \caption{Illustration of different modeling paradigms for TSFMs. (a) Unimodal TSFMs  on numerical time series.
(b) Role-agnostic multimodal TSFMs model endogenous (endo-) and exogenous (exo-) modalities through largely shared mechanisms.
(c) Our role-aware multimodal TSFM separately models endo- and exo- modalities according to their distinct characteristics.
}
  \label{fig: intro}
  \vspace{-8mm}
\end{figure}

Time series forecasting is critical to real-world applications~\citep{sezer2020financial,qiu2024tfb,wu2025k2vae}.
Recent time series foundation models (TSFMs) have demonstrated strong generalization through pretraining on large-scale numerical time series corpora, learning transferable temporal patterns~\citep{ansari2025chronos2,zeus,tirex2}, as illustrated in Figure~\ref{fig: intro}(a).
Building on this progress, recent studies have developed multimodal TSFMs that are pretrained on multimodal data and  designed to jointly model time series with information such as text and images~\citep{wu2026aurora,cctime,ahamed2026stride}. 
These advances have opened up a broader paradigm of multimodal TSFMs for exploiting heterogeneous information in forecasting.

Within this emerging paradigm, prior multimodal time series studies have distinguished multimodal information by its origin into two categories~\citep{wang2026vot,razmadze2026universal}:
1) \textbf{Endogenous modalities}, which are derived from the time series itself and provide alternative views of the same underlying temporal dynamics; and
2) \textbf{Exogenous modalities}, which originate from external sources and provide additional evidence that may influence future evolution. Despite both being forms of multimodal information, their distinct information origins suggest fundamentally different roles in forecasting.

Existing multimodal TSFMs incorporate both endogenous and exogenous modalities through largely shared multimodal mechanisms~\citep{wu2026aurora,ahamed2026stride,chattime}, as illustrated in Figure~\ref{fig: intro}(b).
However, such treatment overlooks a fundamental asymmetry in how endogenous and exogenous modalities relate to forecasting:
endogenous modalities are \emph{intrinsically coupled with the time series and evolve with its temporal dynamics},
whereas exogenous modalities face \emph{limited multimodal pretraining data and diverse modalities in both type and number across forecasting scenarios}. 
Consequently, ignoring this asymmetry in role-agnostic modeling may hinder the learning of temporal evolution and the generalization across exogenous modality configurations.
To bridge this gap, we aim to explicitly model endogenous and exogenous multimodality with role-aware strategies that better realize their distinct forecasting roles, as illustrated in Figure~\ref{fig: intro}(c).
This presents two key challenges:

\textbf{For endogenous modalities, it remains challenging to capture how
endogenous modalities evolve along with the underlying temporal dynamics.}
Capturing the evolution of endogenous modalities helps characterize the underlying temporal dynamics from multiple perspectives.
However, numerical forecasting losses only evaluate the final forecast values, providing
no modality-specific supervision for how endogenous information should evolve
toward the future.
Existing multimodal TSFMs~\citep{chattime,wu2026aurora}
primarily use endogenous modalities to enrich historical context, without
explicitly learning how these modalities evolve over time.

\textbf{For exogenous modalities, limited multimodal pretraining data makes it
challenging to generalize across domains and across various modality types and
numbers.}
First, the same exo-modality may exhibit different forecasting effects across
domains. For example, the same convective weather condition captured by satellite imagery may indicate increasing precipitation in weather forecasting~\citep{park2025precipitation}, while the same condition can reduce road traffic flow in traffic forecasting~\citep{jia2017traffic}.
Second, forecasting scenarios may differ substantially in the types and numbers
of available exo-modalities.
Such cross-scenario variations pose substantial challenges to model generalization.
This challenge is further exacerbated by the scarcity of exo-multimodal pretraining data, making it difficult to learn transferable exogenous knowledge across diverse scenarios.

To address these challenges, we present \textbf{\name}, which adopts role-aware strategies to model endo- and exo-multimodality differently according to their distinct forecasting characteristics.
\textbf{For endogenous modalities}, we introduce \textit{Endo-Multimodal Predictor} and \textit{Endo-Multimodal Supervision} to explicitly learn how endogenous modalities evolve along with the underlying time series during pretraining.
Specifically, \name{} performs patch-wise fusion of endogenous modalities within
the historical window. In the forecast window, instead of predicting only numerical values, the \textit{Endo-Multimodal Predictor} explicitly predicts
each endo-modality at the patch level. 
These predictions are further guided by \textit{Endo-Multimodal Supervision}, which aligns the predicted endogenous representations with those derived from the ground-truth future sequence, thereby explicitly supervising their evolution from history to the future.
\textbf{For exogenous modalities}, rather than relying on pretraining to acquire such generalization capabilities, we design an exo-multimodal retrieval mechanism that enables rapid downstream adaptation to unseen domains, unseen modality types, and various numbers of modalities without updating the TSFM parameters.
To this end, we propose an \textit{Exo-Multimodal Retrieval Enhancer} module and further design a dedicated training strategy, termed \textit{Endo-Modality Proxy Training}, which trains the retrieval module using diverse endo-modalities as proxies during pretraining.
Specifically, the \textit{Exo-Multimodal Retrieval Enhancer} independently retrieves historical cases for each input exo-modality and uses their corresponding future responses as forecasting evidence, enabling adaptation to downstream tasks with various types and numbers of exo-modalities.
The \textit{Endo-Modality Proxy Training} strategy dynamically samples
combinations of readily available endo-modalities to substitute for scarce
exo-modalities during pretraining.

Our main contributions are summarized as follows:
\begin{itemize}[leftmargin=20pt]
\vspace{-3mm}
    \item We introduce a role-aware multimodal TSFM, named \name{}, which treats
endogenous and exogenous modalities with distinct strategies according to their
different forecasting characteristics, thereby enabling more effective multimodal forecasting.

  \item For endogenous modalities, we introduce the \textit{Endo-Multimodal Predictor} and \textit{Endo-Multimodal Supervision} to explicitly model how endogenous modalities evolve along with the underlying time series.
  
    \item For exogenous modalities, we propose an \textit{Exo-Multimodal Retrieval Enhancer} module supported by an \textit{Endo-Modality Proxy Training} strategy, enabling the model to generalize across unseen domains and modality configurations without updating the TSFM parameters.

    \item Extensive experiments demonstrate that \name{} achieves strong performance across forecasting tasks with and without exogenous modalities.
\end{itemize}

\section{Related work}

\subsection{Multimodal Time Series Forecasting}
Recent studies have increasingly explored multimodal information for time series forecasting. One line of work bridges temporal and language representations by adapting pretrained language models to time series, including GPT4TS~\citep{gpt4ts}, TEST~\citep{test}, Time-LLM~\citep{timellm}, CALF~\citep{liu2025calf}, LLM-Mixer~\citep{llmmixer}, and CC-Time~\citep{cctime}, through reprogramming, representation alignment, or cross-modal interaction. Another line explicitly incorporates auxiliary textual information into forecasting. CMIN~\citep{cmin} and Modality-aware Transformer~\citep{modality-aware} jointly model financial sequences with news or textual reports, while GPT4MTS~\citep{jia2024gpt4mts}, Time-MMD~\citep{time-mmd}, TaTS~\citep{tats}, and VoT~\citep{wang2026vot} further explore the collection, alignment, fusion, or reasoning of contextual text with time series. 
Despite this progress, most existing approaches learn multimodal forecasting through task- or dataset-specific adaptation, 
limiting the transfer of multimodal knowledge to unseen domains.

\subsection{Time Series Foundation Models}
Time series foundation models (TSFMs) leverage large-scale pretraining to learn transferable temporal patterns and enable zero-shot generalization across domains. Representative models, including  UniTS~\citep{units}, TimesFM~\citep{timesfm}, 
and ROSE~\citep{wang2025rose}, explore diverse architectures and pretraining objectives for general-purpose forecasting. More recent models further broaden their capabilities: Toto 2.0~\citep{toto2} investigates large-scale model scaling, Chronos-2~\citep{ansari2025chronos2} supports multivariate and covariate-informed forecasting, TiRex-2~\citep{tirex2} enables efficient recurrent and streaming forecasting, and ZEUS~\citep{zeus} extends pretraining toward multiple time series tasks. Recently, ChatTime~\citep{chattime}, STRIDE~\citep{ahamed2026stride} and Aurora~\citep{wu2026aurora} further introduce multimodal pretraining and modeling into TSFMs. However, these models primarily support exogenous textual information and struggle
to generalize to scenarios with various types and numbers of exo-modalities.
More importantly, existing multimodal TSFMs typically adopt role-agnostic modeling,
using  shared mechanisms across heterogeneous modalities without explicitly
modeling endogenous  and exogenous  modalities according to their distinct forecasting characteristics.


\section{\name}

\begin{figure*}[!htbp]
    \centering
\includegraphics[width=1\linewidth]{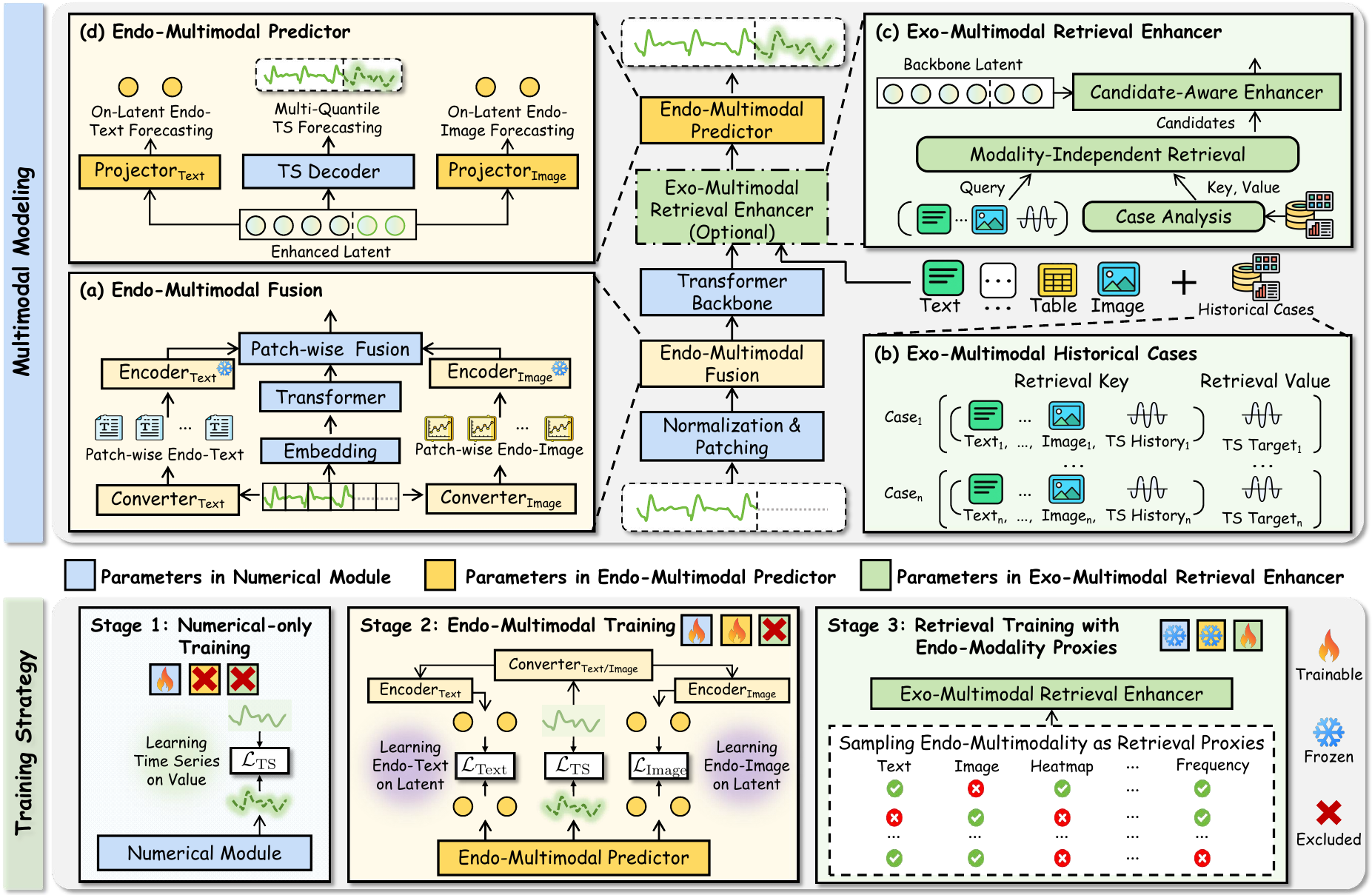}
    \caption{The Multimodal Modeling and Training Strategy  of \name.}
\label{fig: overview}
\vspace{-5mm}
\end{figure*}
In this work, we develop a role-aware multimodal TSFM, named \name{}.
As shown in Figure~\ref{fig: overview}, we design dedicated multimodal modeling
mechanisms and training strategies for endo- and exo-modalities according to
their distinct forecasting characteristics.

In terms of model architecture, after normalization and patching, the input time series is processed through three main modules: 1) \textit{Endo-Multimodal Fusion} constructs patch-level endo-modalities and fuses them within each patch to capture fine-grained temporal dynamics; 
2)
After the Transformer backbone, we introduce an optional
\textit{Exo-Multimodal Retrieval Enhancer}, activated only when exo-modalities are available, which retrieves historical cases with similar exogenous
conditions and uses their future responses as forecasting evidence;
3) \textit{Endo-Multimodal Predictor} predicts both 
time series and endo-modalities for each future patch, providing explicit training signals for learning their temporal evolution from history to the future. The details are presented in Section~\ref{sec: Multimodal Modeling}. 

We further design a three-stage training strategy to fully exploit the capabilities of each module and alleviate the scarcity of exo-multimodal data, as
detailed in Section~\ref{sec: Training Strategy}.

\subsection{Multimodal Modeling}
\label{sec: Multimodal Modeling}
Our method explicitly distinguishes endogenous and exogenous multimodal information and models them separately through dedicated modules to achieve role-aware multimodal modeling. Following the channel-independence strategy~\citep{nie2022time}, we aim to predict
the future $H$ time steps at $Q$ quantile levels,
$\hat{\boldsymbol{Y}}\in\mathbb{R}^{H\times Q}$,
from the historical sequence
$\boldsymbol{X}\in\mathbb{R}^{L}$ of length $L$,
optionally augmented with available exogenous multimodal information.

\subsubsection{Endo-Multimodal Fusion}
\label{sec: Endo-Multimodal Fusion}
To better capture the temporal dynamics of the underlying system from multiple perspectives, we construct endogenous multimodal information for each patch.
To balance efficiency and expressiveness, we use statistical-feature descriptions and line plots of multi-order differences as the endogenous modalities, with
their construction details provided in Appendix~\ref{multimodal_gen}.

\textbf{Temporal Encoding.}
Following existing TSFMs~\citep{toto2,ansari2025chronos2}, we zero-pad the historical
sequence to the forecast horizon, resulting in $\tilde{\boldsymbol{X}}\in \mathbb{R}^{L + H}$  and augment it with a normalized time index
$\boldsymbol{J}\in \mathbb{R}^{L + H}$ and an observation mask $\boldsymbol{M}\in \mathbb{R}^{L + H}$. The resulting sequence is patchified and encoded as follows:
\begin{gather}
\boldsymbol{H}_0
=
\operatorname{MLP}\!\left(
\operatorname{Patch}\!\left(
[\tilde{\boldsymbol{X}};\boldsymbol{J};\boldsymbol{M}]
\right)\right)\in\mathbb{R}^{N_{L+F}\times d},
\quad
\boldsymbol{H}^{\mathrm{TS}}
=
\operatorname{Transformer}(\boldsymbol{H}_0)
\in\mathbb{R}^{N_{L+F}\times d},
\end{gather}
where $\tilde{\boldsymbol{X}}\in\mathbb{R}^{L+H}$ is the zero-padded sequence,
$\boldsymbol{J}
=
\left[
-\frac{L}{C},
-\frac{L-1}{C},
\ldots,
0,
\ldots,
\frac{H-1}{C}
\right]$
denotes the normalized time index, and
$\boldsymbol{M}=[\boldsymbol{1}^{L},\boldsymbol{0}^{H}]$
distinguishes observed and forecast positions.
$N_{L+F}$ denote the numbers of patches in the historical and forecast
windows.

\textbf{Endo-Multimodal Construction \& Encoding.}
For each patch, we construct statistical feature descriptions and multi-order
difference plots as the Endo-Text and Endo-Image, respectively.
A frozen pretrained CLIP~\citep{clip} encodes these modalities, and the resulting
token-level features are average-pooled to obtain patch-level representations
$\boldsymbol{H}^{\mathrm{Text}},\boldsymbol{H}^{\mathrm{Image}}
\in\mathbb{R}^{N\times d_{\mathrm{Clip}}}$.

\textbf{Patch-wise Fusion.}
To bridge the representation-space gap between the pretrained CLIP encoder and the temporal module, we employ multi-stage MLPs to fuse the time-series
$\boldsymbol{H}^{\mathrm{TS}}$, text
$\boldsymbol{H}^{\mathrm{Text}}$, and image
$\boldsymbol{H}^{\mathrm{Image}}$ representations at the patch level, which is formulated as follows:
\begin{gather}
    \boldsymbol{H}_1 =
    \operatorname{MLP}_1
    \left(
    \operatorname{Concat}
    \left(
    \boldsymbol{H}^{\mathrm{TS}},
    \boldsymbol{H}^{\mathrm{Text}}
    \right)
    \right),
    \quad
    \boldsymbol{H}_2 =
    \operatorname{MLP}_2
    \left(
    \operatorname{Concat}
    \left(
    \boldsymbol{H}^{\mathrm{TS}},
    \boldsymbol{H}^{\mathrm{Image}}
    \right)
    \right),
    \\
    \boldsymbol{H} =
    \operatorname{MLP}_3
    \left(
    \boldsymbol{H}_1 +
    \boldsymbol{H}_2 +
    \boldsymbol{H}^{\mathrm{TS}}
    \right),
\end{gather}
where $\operatorname{MLP}_1:=\mathbb{R}^{d+d_\text{Clip}}\to\mathbb{R}^{d}$, $\operatorname{MLP}_2:=\mathbb{R}^{d+d_\text{Clip}}\to\mathbb{R}^{d}$, and
$\operatorname{MLP}_3:=\mathbb{R}^{d}\to\mathbb{R}^{d}$ denote learnable multilayer perceptrons operating
independently on each patch.
$\boldsymbol{H}_1$ and $\boldsymbol{H}_2$ fuse the time-series representation
with the text and image representations, respectively, and are further aggregated
by $\operatorname{MLP}_3$ to obtain the final patch-level representation
$\boldsymbol{H}$.

\subsubsection{Exo-Multimodal Retrieval Enhancer}
\label{sec: Exo-Multimodal Retrieval Enhancer}
\begin{wrapfigure}{r}{0.45\columnwidth}
\vspace{-4mm}
  \centering
  \raisebox{0pt}[\height][\depth]{\includegraphics[width=0.45\columnwidth]{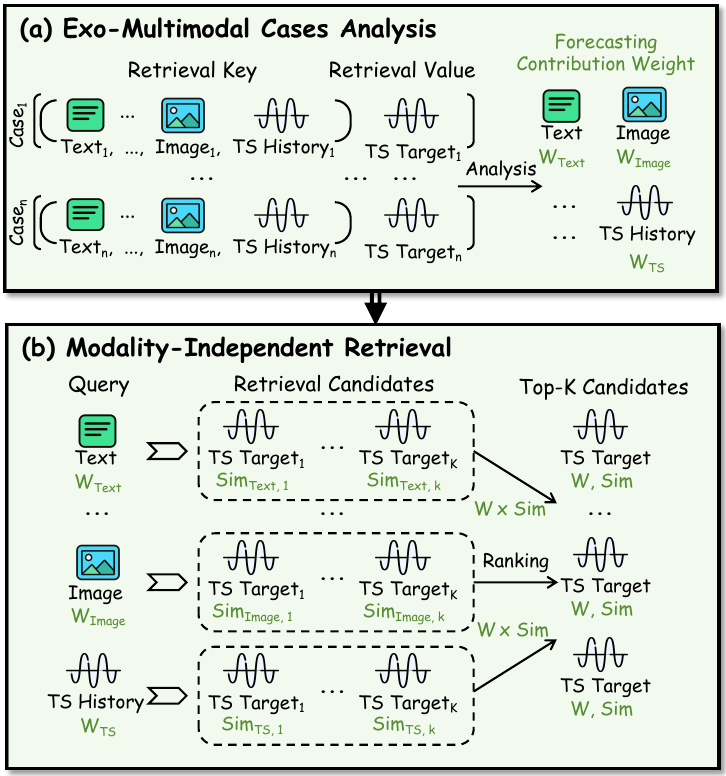}}
  \caption{Key steps of the \textit{Exo-Multimodal Retrieval Enhancer}.}
  \label{fig: retrieval}
  \vspace{-10mm}
\end{wrapfigure}





We introduce this module to accommodate various downstream exogenous modality types and numbers through retrieval from historical cases without updating TSFM parameters. The module is pretrained with an \textit{Endo-Modality Proxy Training} strategy, which dynamically samples combinations of available endo-modalities as retrieval proxies.

To reduce the noise introduced by irrelevant modalities, we first design an \textit{Exo-Multimodal Cases Analysis} module (Figure~\ref{fig: retrieval}(a)), which analyzes the provided historical cases before retrieval and identifies the forecasting contribution of each modality.

To handle various modality types and numbers, we propose
\textit{Modality-Independent Retrieval} (Figure~\ref{fig: retrieval}(b)),
where each modality independently retrieves top-$K$ historical cases as
candidates, which are then merged and re-ranked by similarity scores and modality forecasting contributions.
The selected candidates are incorporated into the forecasting representation
through the \textit{Candidate-Aware Enhancer}.

\textbf{Exo-Multimodal Cases Analysis.}
Given $N_R$ historical cases $\boldsymbol{R}=\{\boldsymbol{r}_i\}_{i=1}^{N_R}$,
each case is represented as
$
\boldsymbol{r}_i=
(\boldsymbol{H}^{\mathrm{his}}_i,
\{\boldsymbol{E}^{m}_i\}_{m\in\mathcal{M}},
\boldsymbol{Y}^{\mathrm{fut}}_i),
$
where $\boldsymbol{H}^{\mathrm{his}}_i$ and
$\boldsymbol{Y}^{\mathrm{fut}}_i$ denote its historical time-series representation
and future sequence, respectively.
$\mathcal{M}$ denotes the set of available exo-modalities with
$M=|\mathcal{M}|$, and $\boldsymbol{E}^{m}_i$ denotes the representation of
modality $m\in\mathcal{M}$.

As illustrated in Figure~\ref{fig: retrieval}(a), we estimate the
\textit{Forecasting Contribution Weight}
$\boldsymbol{W}=(W^1,\ldots,W^M)\in\mathbb{R}^{M}$
by evaluating how effectively each modality retrieves historical cases whose
future responses are informative for forecasting.

Specifically, for each reference case $\boldsymbol{r}_i$ and modality
$m\in\mathcal{M}$, we retrieve the top-$K$ other reference cases according to
the modality-specific similarity
$s_{ij}^m=s(\boldsymbol{E}_i^m,\boldsymbol{E}_j^m)$
and aggregate their future sequences into a retrieval-based future estimate:
\begin{gather}
\mathcal{N}_i^m
=
\operatorname{TopK}_{j\neq i}(s_{ij}^m),
\quad
\hat{\boldsymbol{Y}}_i^m
=
\sum_{j\in\mathcal{N}_i^m}
\operatorname{Softmax}_{j\in\mathcal{N}_i^m}(s_{ij}^m)
\boldsymbol{Y}_j^{\mathrm{fut}},
\\
e_i^m
=
\operatorname{MSE}
\left(
\hat{\boldsymbol{Y}}_i^m,
\boldsymbol{Y}_i^{\mathrm{fut}}
\right),
\quad
w_i^m
=
\operatorname{Softmax}_{m\in\mathcal{M}}(-e_i^m), \quad
W^m
=
\frac{1}{N_R}
\sum_{i=1}^{N_R} w_i^m .
\end{gather}
Thus, a larger $W^m$ indicates higher retrieval-based forecasting utility of modality $m$.

\textbf{Modality-Independent Retrieval.}
As illustrated in Figure~\ref{fig: retrieval}(b), given a query $q$, we
perform retrieval independently for each available modality to accommodate
various modality types and numbers. Each modality retrieves its top-$K$
reference cases, which are then merged and re-ranked using the
\textit{Forecasting Contribution Weight}.
Let
$s_{qj}^m=s(\boldsymbol{E}_q^m,\boldsymbol{E}_j^m)$
denote the similarity between the query and reference case $\boldsymbol{r}_j$
under modality $m$:
\begin{gather}
\mathcal{C}_q^m
=
\operatorname{TopK}_{j}(s_{qj}^m),
\quad
\mathcal{U}_q
=
\bigcup_{m\in\mathcal{M}}\mathcal{C}_q^m,
\quad
S_{qj}
=
\sum_{m\in\mathcal{M}}
W^m s_{qj}^m,
\quad
\mathcal{C}_q
=
\operatorname{TopK}_{j\in\mathcal{U}_q}(S_{qj}),
\end{gather}
where $\mathcal{C}_q^m$ denotes the top-$K$ reference cases retrieved by
modality $m$, $\mathcal{U}_q$ denotes the union of all modality-specific
candidate sets, $S_{qj}$ denotes the contribution-weighted retrieval score of
reference case $\boldsymbol{r}_j$ for query $q$, and $\mathcal{C}_q$ denotes
the final top-$K$ reference cases after re-ranking.

\textbf{Candidate-Aware Enhancer.}
Given the final candidate set $\mathcal{C}_q$, we incorporate the retrieved
information into the backbone representation $\boldsymbol{H}^B$ through cross-attention.
For each candidate $\boldsymbol{r}_j\in\mathcal{C}_q$, we encode its
representation $\boldsymbol{H}^{\mathrm{cand}}_j$ and construct a candidate identifier $\boldsymbol{I}_{qj}$ from its modality
contribution $W^m$ and retrieval similarity $s_{qj}^m$:
\begin{gather}
\boldsymbol{H}^{\mathrm{cand}}_j
=
\operatorname{MLP}_{\mathrm{cand}}(\boldsymbol{r}_j),
\quad
\boldsymbol{I}_{qj}
=
\operatorname{MLP}_{I}
\left(
[W^m,\,s_{qj}^m]
\right),
\quad
\boldsymbol{Q}
=
\operatorname{MLP}_{Q}(\boldsymbol{H}^{B}),
\\
\boldsymbol{K}_j,\boldsymbol{V}_j
=
\operatorname{MLP}_{KV}
\left(
\boldsymbol{H}^{\mathrm{cand}}_j+\boldsymbol{I}_{qj}
\right), \quad
\boldsymbol{H}^{\mathrm{Enh}}
=
\operatorname{CrossAttn}
(\boldsymbol{Q},\boldsymbol{K},\boldsymbol{V}),
\end{gather}
where $\boldsymbol{K}$ and $\boldsymbol{V}$ are obtained by stacking the
candidate-wise representations $\boldsymbol{K}_j$ and $\boldsymbol{V}_j$
over all $\boldsymbol{r}_j\in\mathcal{C}_q$, respectively. $\boldsymbol{H}^{\mathrm{Enh}}$ is the enhanced representation of the input time series.

\subsubsection{Endo-Multimodal Predictor}

To explicitly model the evolution of endogenous multimodal information from
history to the future, we jointly predict the future time series together with its Endo-Text and Endo-Image modalities.
Rather than directly generating text tokens or image pixels, we predict their
representations for computational efficiency and to avoid unnecessary
token- and pixel-level generation noise.

After optionally incorporating exogenous information through the
\textit{Exo-Multimodal Retrieval Enhancer}, we retain only the forecast-window
representations
$\boldsymbol{H}^{\mathrm{Enh}}_{\mathrm{pred}}
\in\mathbb{R}^{N_f\times d}$
for efficient decoding.

\textbf{Time Series Decoding.}
To support uncertainty quantification, we employ an MLP-based multi-quantile prediction head to produce probabilistic
forecasts at $Q$ quantile levels:
\begin{gather}
\hat{\boldsymbol{Y}}
=
\operatorname{QuantileHead}
\left(
\boldsymbol{H}^{\mathrm{Enh}}_{\mathrm{pred}}
\right)
\in\mathbb{R}^{H\times Q}.
\end{gather}



\textbf{Endo-Multimodal Projector.}
We further project the forecast-window representations into the CLIP space to
predict future Endo-Text representations  $\boldsymbol{F}^{\mathrm{Text}}$ and Endo-Image representations $\boldsymbol{F}^{\mathrm{Image}}$:
\begin{gather}
\boldsymbol{F}^{\mathrm{Text}}
=
\operatorname{Project}_{\mathrm{T}}
\left(
\boldsymbol{H}^{\mathrm{Enh}}_{\mathrm{pred}}
\right)
\in\mathbb{R}^{N_f\times d_{\mathrm{Clip}}},~
\boldsymbol{F}^{\mathrm{Image}}
=
\operatorname{Project}_{\mathrm{Image}}
\left(
\boldsymbol{H}^{\mathrm{Enh}}_{\mathrm{pred}}
\right)
\in\mathbb{R}^{N_f\times d_{\mathrm{Clip}}}.
\end{gather}
Here, the projectors are MLPs that map the forecast representations to the CLIP space.

\subsection{Training Strategy}
\label{sec: Training Strategy}
To progressively equip \name{} with numerical forecasting, endogenous multimodal
modeling, and exogenous retrieval capabilities, we adopt a three-stage training
strategy, as illustrated in Figure~\ref{fig: overview}:
1) \textit{Numerical-only Training} establishes the basic temporal modeling and
forecasting capability;
2) \textit{Endo-Multimodal Training} further learns endogenous multimodal
representations and their evolution from history to the future; and
3) \textit{Retrieval Training with Endo-Modality Proxies} trains the
\textit{Exo-Multimodal Retrieval Enhancer} using readily available endogenous
modalities as retrieval proxies, avoiding reliance on large-scale
exo-multimodal pretraining data.


\subsubsection{Numerical-only Training}
In this stage, we train only the numerical forecasting pathway, including the
temporal encoder, \textit{Transformer backbone}, and multi-quantile prediction
head, while all multimodal modules are excluded.
Given the predicted quantile forecasts
$\hat{\boldsymbol{Y}}\in\mathbb{R}^{H\times Q}$ and the ground-truth future
sequence $\boldsymbol{Y}\in\mathbb{R}^{H}$, we optimize these parameters using
the Pinball loss~\citep{ansari2025chronos2}:
\begin{equation}
\mathcal{L}_{\mathrm{TS}}
=
\frac{1}{HQ}
\sum_{h=1}^{H}
\sum_{q=1}^{Q}
\max\left(
\tau_q\left(Y_h-\hat{Y}_{h,q}\right),
(\tau_q-1)\left(Y_h-\hat{Y}_{h,q}\right)
\right),
\end{equation}
where $\tau_q\in(0,1)$ denotes the $q$-th quantile level.


\subsubsection{Endo-Multimodal Training}
In this stage, we optimize the numerical pathway and endogenous multimodal modules, while excluding the \textit{Exo-Multimodal Retrieval Enhancer} and freezing the pretrained CLIP encoder.

To provide explicit supervision for endogenous evolution, we transform the
ground-truth future sequence $\boldsymbol{Y}$ into Endo-Text and Endo-Image
modalities using the same procedure as in
Section~\ref{sec: Endo-Multimodal Fusion}, and encode them with the frozen CLIP
encoder to obtain future representation targets
$\boldsymbol{T}^{\mathrm{Text}},\boldsymbol{T}^{\mathrm{Image}}
\in\mathbb{R}^{N_f\times d_{\mathrm{Cli'p}}}$.
The representation-level supervision is defined as follows:
\begin{gather}
\mathcal{L}_{\mathrm{Text}}
=
\frac{
\left\|
\boldsymbol{F}^{\mathrm{Text}}
-
\boldsymbol{T}^{\mathrm{Text}}
\right\|_F^2
}{
N_f d_{\mathrm{Clip}}
},
\qquad
\mathcal{L}_{\mathrm{Image}}
=
\frac{
\left\|
\boldsymbol{F}^{\mathrm{Image}}
-
\boldsymbol{T}^{\mathrm{Image}}
\right\|_F^2
}{
N_f d_{\mathrm{Clip}}
},
\end{gather}  where $\|\cdot\|_F$ denotes the Frobenius norm.
The overall objective is
$
\mathcal{L}_\text{Multi}
=
\mathcal{L}_{\mathrm{TS}}
+
\lambda_{\mathrm{Text}}\mathcal{L}_{\mathrm{Text}}
+
\lambda_{\mathrm{Image}}\mathcal{L}_{\mathrm{Image}},
$
where $\lambda_{\mathrm{Text}}$ and $\lambda_{\mathrm{Image}}$ balance the
two endogenous supervision terms.

\subsubsection{Retrieval Training with Endo-Modality Proxies}
In this stage, we optimize only the \textit{Exo-Multimodal Retrieval Enhancer} and forecasting head.
We introduce \textit{Endo-Modality Proxy Training}, which dynamically samples various types and numbers of endo-modalities as retrieval proxies, 
avoiding the need for exo-multimodal pretraining data.

Let $\mathcal{M}_{\mathrm{endo}}$ denote the set of available endogenous
modalities. For each training query $q$, we randomly sample a non-empty subset
$\mathcal{S}_q$ and treat the selected modalities as retrieval proxies.

The sampled proxies are then processed by the same retrieval pipeline described
in Section~\ref{sec: Exo-Multimodal Retrieval Enhancer}, to obtain the final enhanced latent $\boldsymbol{H}^\text{Enh}_\text{proxy}$.
The forecasting head then generates the prediction
$\hat{\boldsymbol{Y}}^\text{proxy}$, and only the parameters of the retrieval enhancer and head are
optimized using the forecasting loss:
\begin{equation}
\mathcal{L}_{\mathrm{proxy}}
=
\frac{1}{HQ}
\sum_{h=1}^{H}
\sum_{q=1}^{Q}
\max\left(
\tau_q\left(Y_h-\hat{Y}_{h,q}^\text{proxy}\right),
(\tau_q-1)\left(Y_h-\hat{Y}_{h,q}^\text{proxy}\right)
\right).
\end{equation}
The endogenous proxies are not intended to mimic the semantic content of
exogenous modalities; instead, they expose the retrieval enhancer to various
modality types, numbers, and combinations while preserving the same
retrieval-to-response learning process used at inference time.

\section{Experiments}
\begin{wrapfigure}{r}{0.33\columnwidth}
\vspace{-8mm}
  \centering
  \raisebox{0pt}[\height][\depth]{\includegraphics[width=0.33\columnwidth]{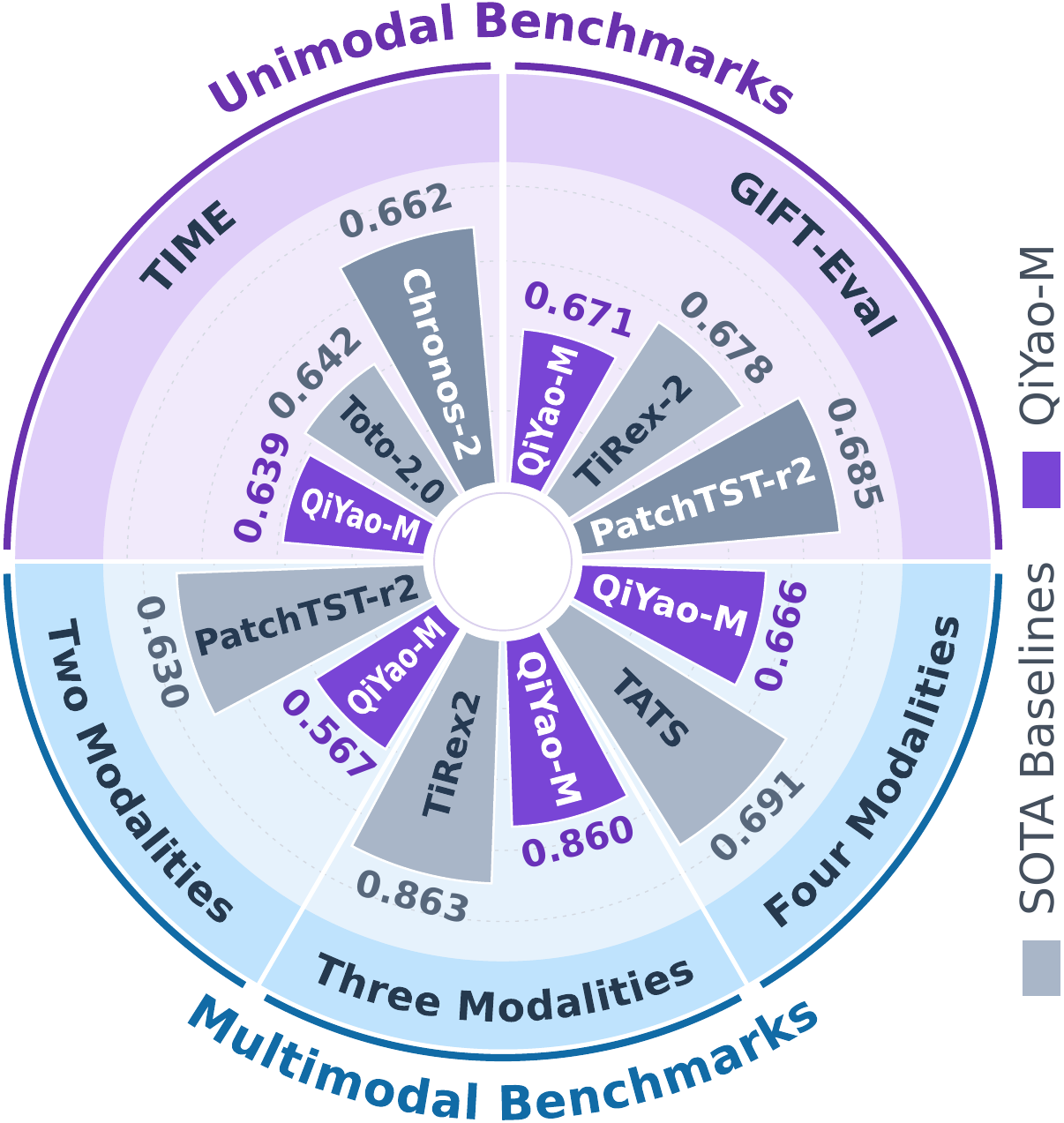}}
  \caption{Evaluation summary.}
  \label{fig: evaluation}
\end{wrapfigure} 

We conduct extensive experiments to evaluate \name{}. Detailed experimental settings are provided in Appendix~\ref{app: exp details}, including the pretraining corpus (Appendix~\ref{app: pretraining corpus}), benchmarks (Appendix~\ref{app: benchmarks}), baselines (Appendix~\ref{app: baselines}), and other implementation details. Sections~\ref{sec: unimodal forecasting} and~\ref{sec: multimodal forecasting} evaluate forecasting performance on unimodal benchmarks (GIFT-Eval~\citep{aksu2024gift} and TIME~\citep{qiao2026s}) and multimodal benchmarks with 2--4 modalities (Time-MMD~\citep{time-mmd}, MoTime~\citep{zhou2025motime}, and FinMultiTime~\citep{xu2025finmultitime}), respectively. Section~\ref{sec: model analysis} provides further model analyses. 

As an overview, Figure~\ref{fig: evaluation} shows that \name{} achieves state-of-the-art MASE on unimodal benchmarks and MSE on multimodal benchmarks, demonstrating strong performance in scenarios both with and without exo-modalities.




\subsection{Performance on Unimodal Benchmarks}
\label{sec: unimodal forecasting}

\begin{figure*}[!h]
  \centering
  \begin{minipage}{0.48\linewidth}
    \centering
    \includegraphics[width=\linewidth]{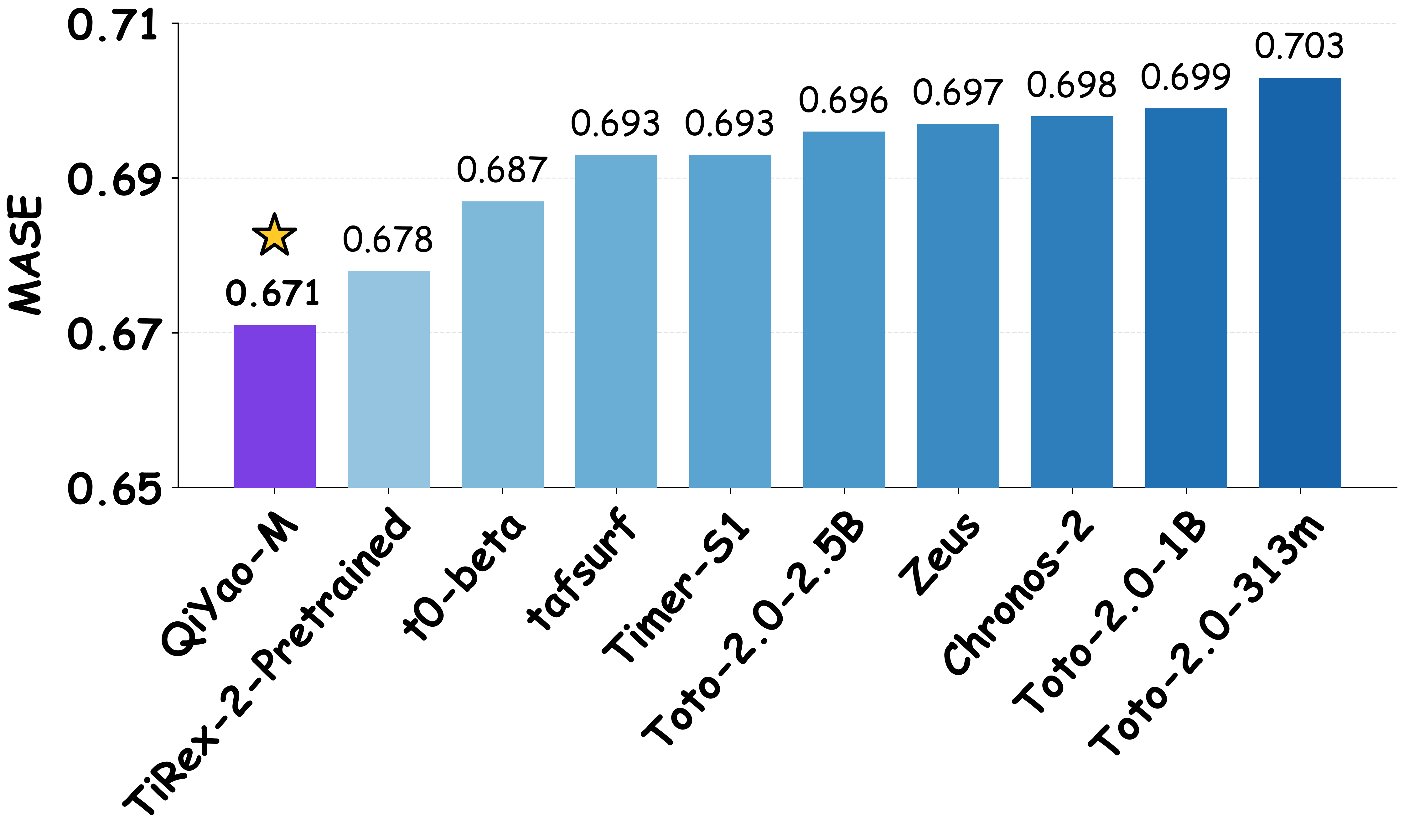}\\[-1mm]
    {\scriptsize (a) MASE result on GIFT-Eval}
  \end{minipage}\hfill
  \begin{minipage}{0.48\linewidth}
    \centering
    \includegraphics[width=\linewidth]{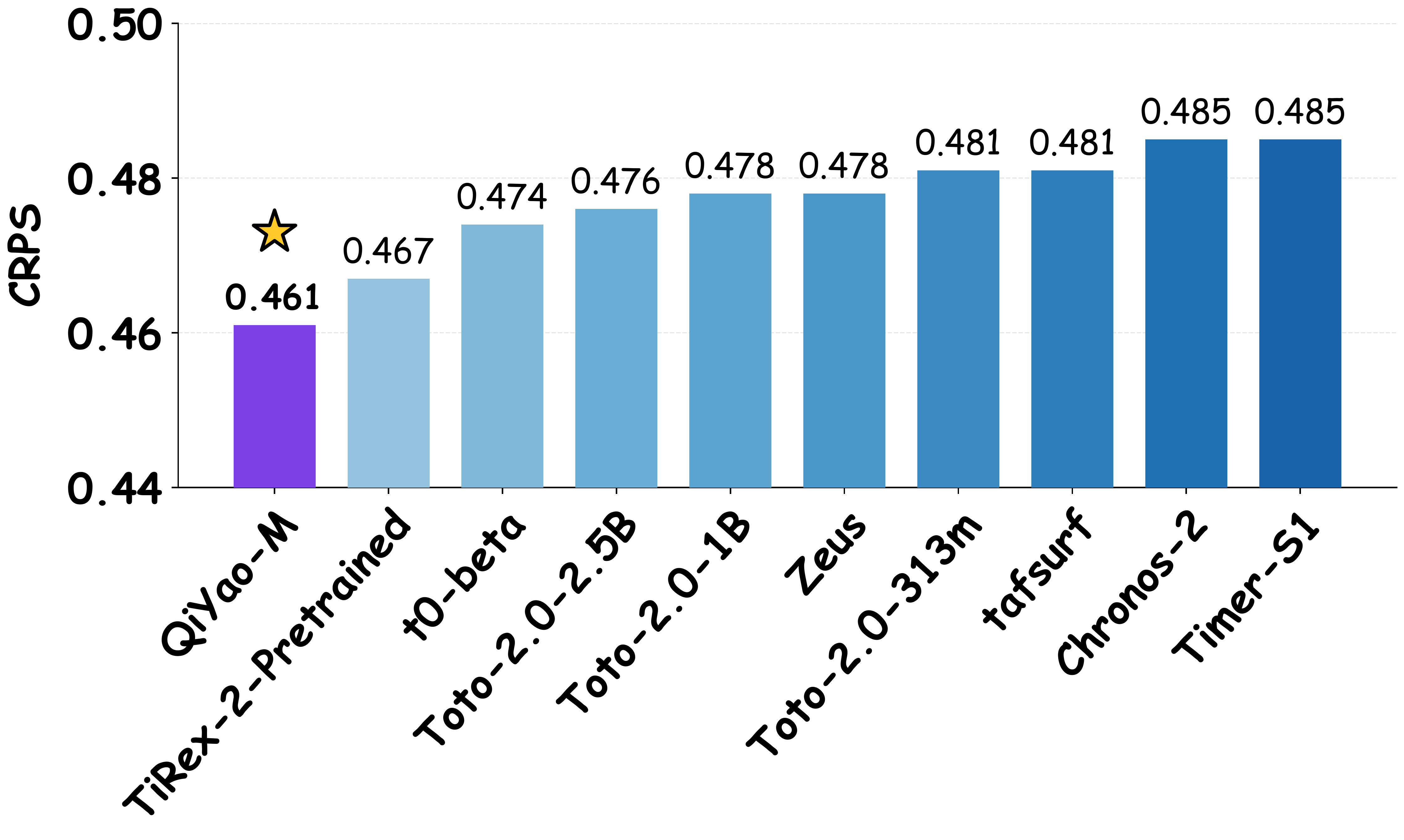}\\[-1mm]
    {\scriptsize (b) CRPS result on GIFT-Eval}
  \end{minipage}

  \vspace{1mm}

  \begin{minipage}{0.48\linewidth}
    \centering
    \includegraphics[width=\linewidth]{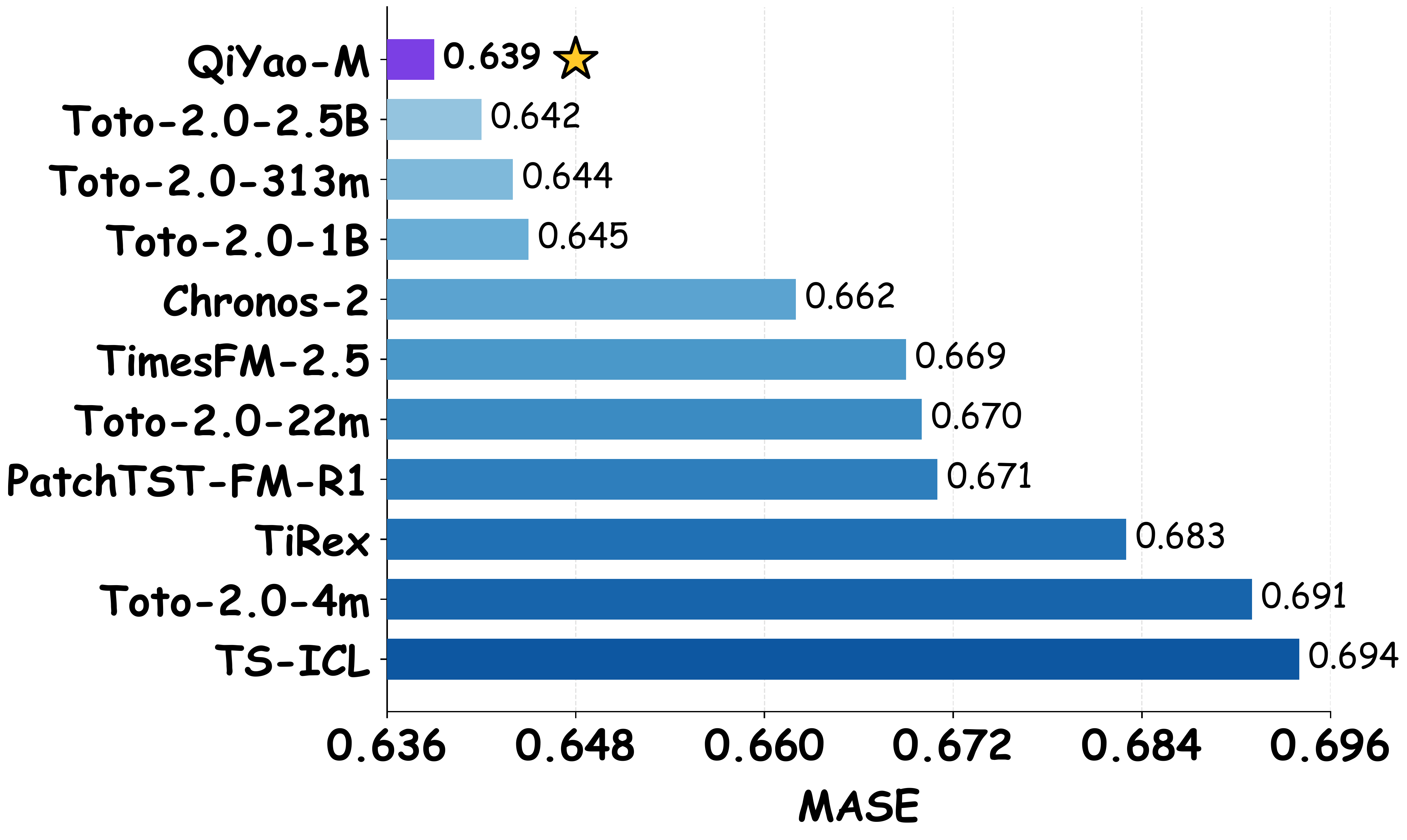}\\[-1mm]
    {\scriptsize (c) MASE result on TIME }
  \end{minipage}\hfill
  \begin{minipage}{0.48\linewidth}
    \centering
    \includegraphics[width=\linewidth]{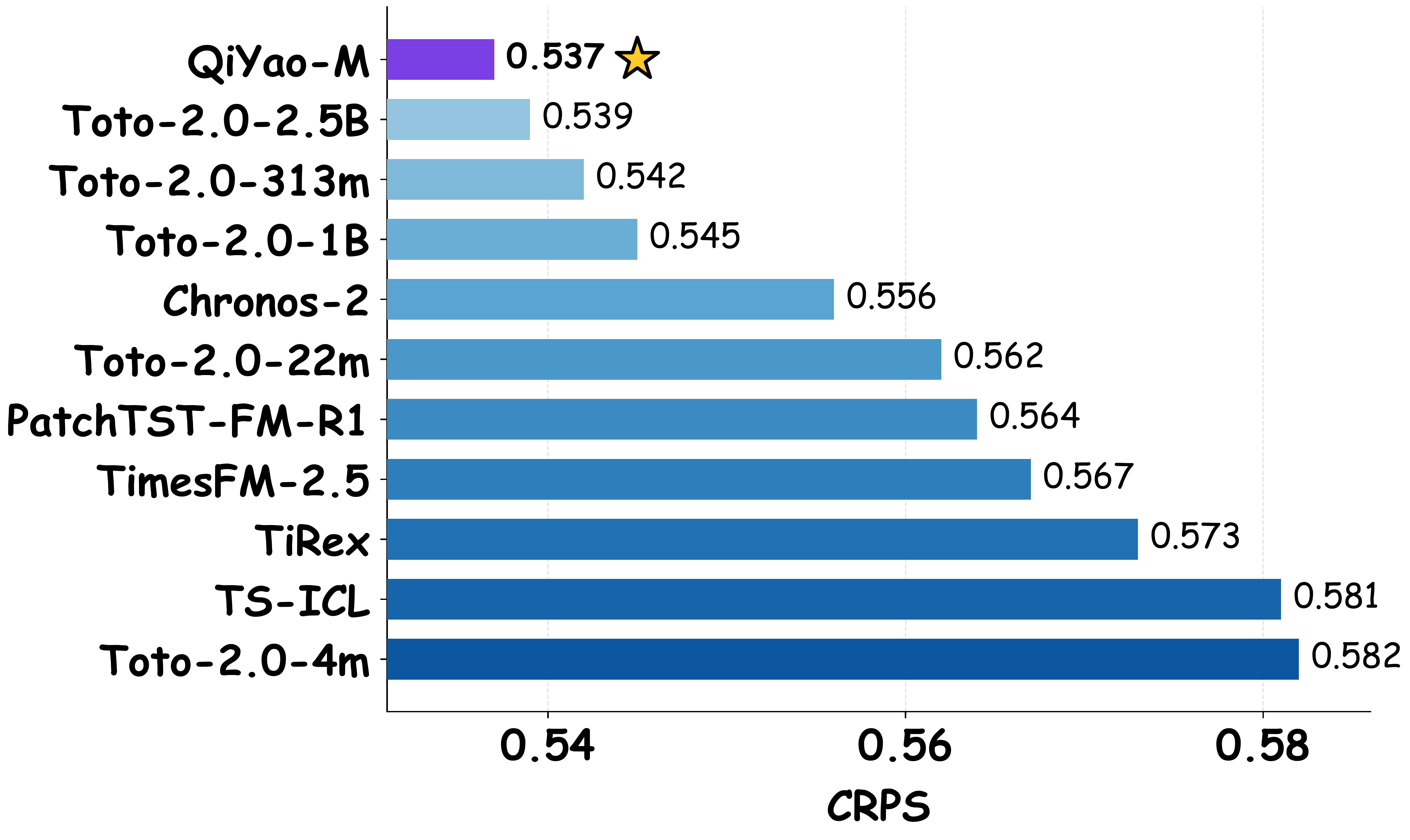}\\[-1mm]
    {\scriptsize (d) CRPS result on TIME}
  \end{minipage}

  \caption{Performance of \name{} on unimodal benchmarks.}
  \label{fig:GIFTEVAL}
\end{figure*}

To evaluate \name{} with endo-modalities only, we conduct experiments on the unimodal benchmarks GIFT-Eval and TIME. As shown in Figure~\ref{fig:GIFTEVAL}, \name{} achieves the best MASE and CRPS on both benchmarks. 
On GIFT-Eval, \name{} reduces MASE and CRPS by 1.0\% and 1.3\%, respectively, compared with the strongest baseline TiRex-2-Pretrained. It also achieves consistently strong MASE and CRPS rankings across datasets (see Appendix~\ref{app: GIFT-Eval Rankings}), demonstrating robust generalization across diverse forecasting scenarios.
On TIME, \name{} achieves a MASE of 0.639 and a CRPS of 0.537, outperforming Toto-2.0-2.5B at 0.642 and 0.539. These consistent gains demonstrate the effectiveness of endo-multimodal modeling across different forecasting benchmarks.

\subsection{Performance on  Multimodal benchmarks}
\label{sec: multimodal forecasting}
To evaluate \name{} with both endo- and exo-modalities, we conduct experiments on multimodal datasets with diverse modality combinations from Time-MMD, MoTime, and FinMultiTime. As shown in Table~\ref{tab:multimodal_results}, incorporating exo-modalities reduces the average MSE of \name{} by 5.1\%. Compared with baseline foundation and end-to-end models, \name{} ranks first on 14/18 metrics and within the top two on 17/18 metrics. Compared with the strongest unimodal TSFM on each dataset, \name{} reduces the MSE by 4.3\% on average. Moreover, while existing multimodal TSFMs primarily support exogenous textual information and struggle with various modality types and numbers, \name{} effectively handles diverse exo-modalities, achieving an average MSE reduction of 14.6\%. These consistent improvements demonstrate the effectiveness of \name{} in utilizing exogenous information for forecasting across different domains.

\begin{table*}[!htbp]
\label{tab:multimodal_results}
\centering
\caption{Average forecasting results on multimodal datasets. 
The best and second-best results are highlighted in \rkw{purple} 
and \rjw{blue}, respectively. Full results are listed in Section~\ref{app: Full Results} of Appendix~\ref{app: More Results}.}
\label{tab:multimodal_results}
\renewcommand{\arraystretch}{0.95}
\setlength{\tabcolsep}{3pt}
\resizebox{\textwidth}{!}{
\begin{tabular}{
l|
cc|cc|cc|cc|cc|
cc|cc|
cc|cc
}
\toprule
\rowcg
\textbf{Modal Type}
& \multicolumn{10}{c|}{\textbf{TS + Text}}
& \multicolumn{4}{c|}{\textbf{TS + Text + Image}}
& \multicolumn{4}{c}{\textbf{TS + Text + Image + Table}} \\ 
\midrule 

\multirow{2}{*}{\textbf{Methods}} 
& \multicolumn{2}{c|}{\textbf{Agriculture}}
& \multicolumn{2}{c|}{\textbf{Climate}}
& \multicolumn{2}{c|}{\textbf{Energy}}
& \multicolumn{2}{c|}{\textbf{Health}}
& \multicolumn{2}{c|}{\textbf{Social Good}}
& \multicolumn{2}{c|}{\textbf{TAOBAO}}
& \multicolumn{2}{c|}{\textbf{Tianchi}}
& \multicolumn{2}{c|}{\textbf{HS300}}
& \multicolumn{2}{c}{\textbf{SP500}} \\

& MSE & MAE
& MSE & MAE
& MSE & MAE
& MSE & MAE
& MSE & MAE
& MSE & MAE
& MSE & MAE
& MSE & MAE
& MSE & MAE \\

\midrule

\rowcg\multicolumn{19}{c}{\textit{\textbf{Multimodal End-to-End Models}}} \\[-1pt]
\midrule

\methodlogo{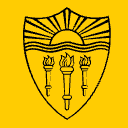}{GPT4MTS}
& 0.225 & 0.298
& 1.182 & 0.889
& 0.262 & 0.380
& 1.464 & 0.799
& 0.920 & 0.450
& 0.534 & 0.257
& 1.320 & 0.147
& 0.780 & 0.535
& 0.767 & 0.642 \\

\methodlogo{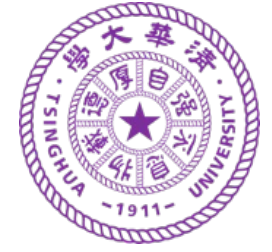}{CALF}
& 0.250 & 0.315
& 1.286 & 0.922
& 0.244 & 0.365
& 1.491 & 0.775
& 0.906 & 0.401
& 0.512 & 0.250
& 1.320 & 0.119
& 0.674 & 0.502
& 0.743 & 0.617 \\

\methodlogo{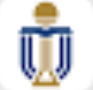}{Time-VLM}
& 0.237 & 0.302
& 1.195 & 0.899
& 0.260 & 0.374
& 1.565 & 0.860
& 0.868 & 0.444
& 0.528 & 0.271
& 1.368 & 0.137
& \best{0.662} & 0.499
& 0.756 & 0.618 \\

\methodlogo{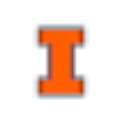}{TATS}
& 0.215 & 0.301
& 1.180 & 0.887
& 0.255 & 0.368
& 1.356 & 0.767
& 0.918 & 0.428
& 0.525 & 0.263
& 1.374 & 0.136
& 0.717 & 0.509
& \second{0.665} & 0.581 \\

\midrule
\rowcg\multicolumn{19}{c}{\textit{\textbf{Unimodal Foundation Models}}} \\[-1pt]
\midrule
\methodlogo{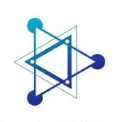}{Zeus}
& 0.203 & 0.295
& 0.852 & 0.738
& 0.239 & 0.337
& 1.590 & 0.845
& 0.999 & 0.434
& \second{0.425} & \second{0.172}
& 1.378 & 0.083
& 0.755 & 0.524
& 0.858 & 0.628 \\

\methodlogo{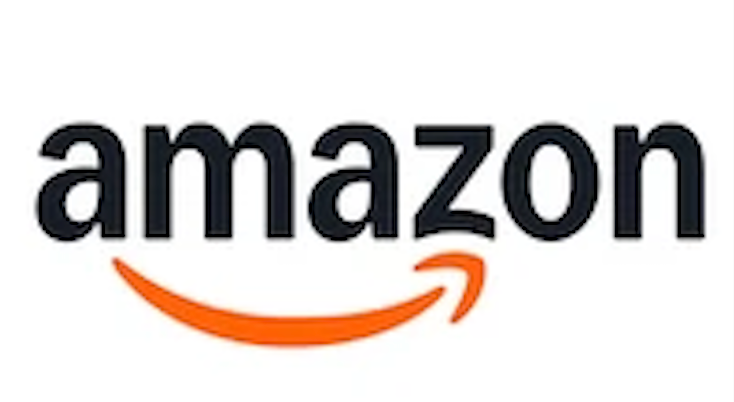}{Chronos-2}
& 0.238 & 0.319
& 0.859 & 0.730
& 0.226 & \second{0.326}
& 1.054 & 0.663
& 0.902 & 0.384
& 0.445 & 0.184
& 1.453 & 0.083
& 0.810 & 0.501
& 0.717 & 0.577 \\

\methodlogo{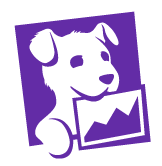}{Toto-2}
& 0.236 & 0.311
& 0.853 & 0.729
& 0.231 & 0.337
& 1.109 & 0.662
& 0.799 & 0.309
& 0.439 & 0.182
& 41.618 & 0.185
& 0.885 & 0.507
& 0.703 & 0.571 \\

\methodlogo{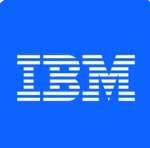}{PatchTST-r2}
& 0.235 & 0.354
& 0.845 & \second{0.724}
& 0.245 & 0.353
& 0.989 & 0.647
& 0.834 & 0.366
& 0.526 & 0.200
& 1.411 & 0.091
& 0.819 & 0.521
& 1.080 & 0.634 \\

\methodlogo{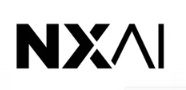}{TiRex2}
& 0.255 & 0.318
& 0.847 & \second{0.724}
& \second{0.224} & 0.330
& 1.097 & 0.638
& \second{0.749} & 0.325
& 0.427 & \best{0.171}
& \best{1.299} & \best{0.074}
& 0.674 & 0.499
& 0.709 & 0.580 \\
\midrule
\rowcg\multicolumn{19}{c}{\textit{\textbf{Multimodal Foundation Models}}} \\
\midrule

\methodlogo{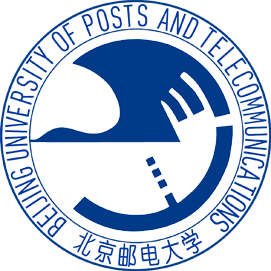}{ChatTime}
& \second{0.196} & 0.293
& 1.144 & 0.856
& 0.258 & 0.355
& 2.278 & 1.001
& 1.319 & 0.541
& 0.479 & 0.201
& 1.676 & 0.087
& 1.208 & 0.686
& 5.985 & 1.779 \\

\methodlogo{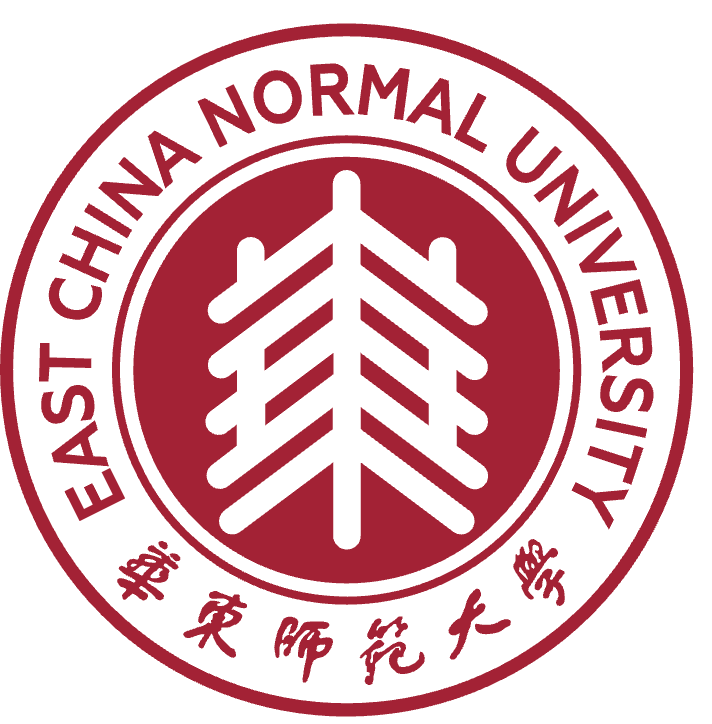}{Aurora}
& 0.272 & 0.348
& 0.865 & 0.749
& 0.255 & 0.370
& 1.553 & 0.850
& 0.838 & 0.516
& 0.442 & 0.210
& 1.403 & 0.090
& 0.736 & 0.524
& 0.862 & 0.649 \\

\midrule

\textbf{Ours} (w/o Exo.)
& 0.199 & \second{0.288}
& \second{0.828} & 0.725
& 0.237 & 0.338
& \second{0.915} & \second{0.612}
& 0.760 & \second{0.296}
& \best{0.417} & \best{0.171}
& 1.423 & 0.081
& 0.690 & 0.496
& 0.694 & \second{0.569} \\

\textbf{Ours} (w Exo.)
& \best{0.174} & \best{0.284}
& \best{0.827} & \best{0.723}
& \best{0.211} & \best{0.325}
& \best{0.901} & \best{0.608}
& \best{0.722} & \best{0.293}
& \best{0.417} & {0.174}
& \second{1.302} & \second{0.076}
& \second{0.672} & \best{0.493}
& \best{0.660} & \best{0.561} \\

\bottomrule
\end{tabular}
}
\vspace{-5mm}
\end{table*}

\subsection{Model Analysis}
\label{sec: model analysis}
\begin{wrapfigure}{r}{0.59\columnwidth}
    \centering

    \captionof{table}{Ablation study of the key components.
    }
    \label{tab:ablation}
    
    \setlength{\tabcolsep}{4pt}
    \renewcommand{\arraystretch}{1.12}
    \resizebox{\linewidth}{!}{%
    \begin{tabular}{c|cc|cc|cc|cc}
    \toprule \rowcg
    & \multicolumn{2}{c|}{\textbf{Endo-Modeling}}
    & \multicolumn{2}{c|}{\textbf{Exo-Modeling}}
    & \multicolumn{2}{c|}{\textbf{Unimodal Bench.}} & \multicolumn{2}{c}{\textbf{Multimodal Bench.}} \\
    \midrule
    \textbf{Row}
    & \makecell{\textbf{Endo.}\\\textbf{Fusion}}
    & \makecell{\textbf{Endo.}\\\textbf{Sup.}}
    & \makecell{\textbf{Retrieval}\\\textbf{Enhancer}}
    & \makecell{\textbf{Case}\\\textbf{Analysis}}
    & \textbf{GIFT-Eval}
    & \textbf{TIME}
    & \makecell{\textbf{FinMulti}\\\textbf{Time}}
    & \makecell{\textbf{Time}\\\textbf{-MMD}} \\
    \midrule \rowcg
    1 & \xmark & \xmark & \xmark & \xmark
      & 0.686 & 0.646 & 0.741 & 0.605 \\ \midrule
    2 & \cmark & \xmark & \xmark & \xmark
      & 0.678 & 0.644 & 0.728 & 0.591 \\ \rowcg
    3 & \cmark & \cmark & \xmark & \xmark
      & \textbf{0.671} & \textbf{0.639} & 0.712 & 0.588 \\ \midrule
    4 & \cmark & \cmark & \cmark & \xmark
      & -- & -- & 0.672 & 0.573 \\ \rowcg
    5 & \cmark & \cmark & \cmark & \cmark
      & -- & -- & \textbf{0.666} & \textbf{0.567} \\
    \bottomrule
    \end{tabular}%
    }

    {\scriptsize\raggedright
    Exo-Modeling does not affect unimodal benchmarks; thus, the
    corresponding entries are marked as ``--''.\par
    }
    \vspace{2mm}
    \includegraphics[
        width=\linewidth
    ]{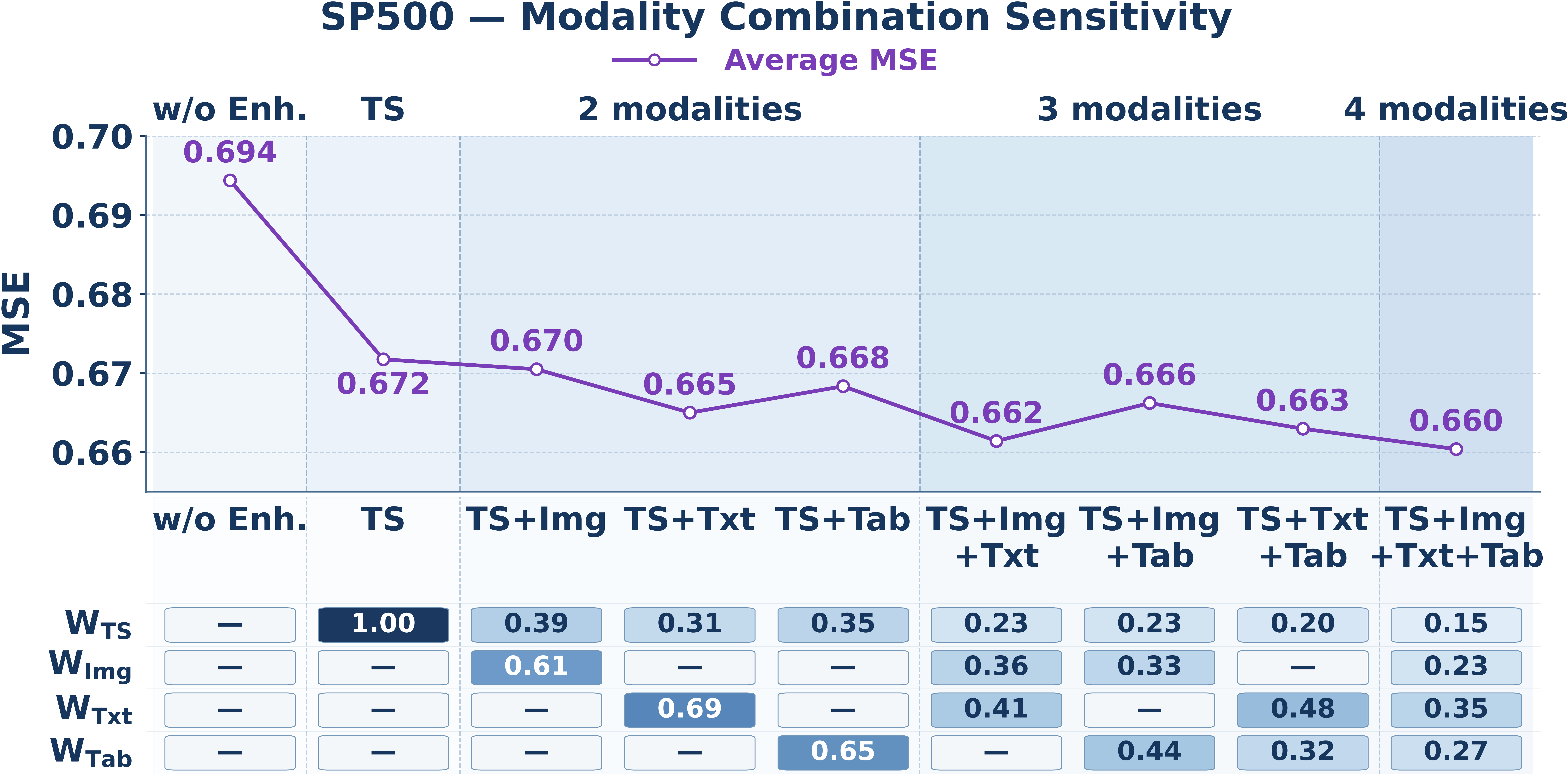}

    \captionof{figure}{Modality analysis.}
    \label{fig:modality_analysis}
    \vspace{-4mm}
\end{wrapfigure}

\textbf{Ablation Studies.}
To analyze the contribution of each component of \name{}, we conduct ablation studies on four benchmarks and report MASE or MSE results in Table~\ref{tab:ablation}.
We have the following observations: 1) Comparing Rows 1 and 2, \textit{Endo-Multimodal Fusion} \textit{(Endo. Fusion)} consistently improves performance, showing the benefit of incorporating patch-level endo-multimodal information. 2) Comparing Rows 2 and 3, \textit{Endo-Multimodal Predictor} and \textit{Endo-Multimodal Supervision} \textit{(Endo. Sup.)} further improve performance by explicitly supervising the evolution of endo-modalities. 3) Comparing Rows 3 and 4, \textit{Exo-Multimodal Retrieval Enhancer} \textit{(Retrieval Enhancer)} 
brings further improvements by incorporating exogenous information from historical reference cases. 4) Comparing Rows 4 and 5, \textit{Exo-Multimodal Cases Analysis (Cases Analysis)}  further improves performance by estimating the \textit{Forecasting Contribution Weight} of different exo-modalities. Overall, the complete \name{} achieves the best performance compared to all the variants across all benchmarks.

\textbf{Modality Analysis.}
Figure~\ref{fig:modality_analysis} examines different modality combinations on SP500. We  have the following observations: 1) Adding exogenous modalities generally improves forecasting performance, with consistent gains over TS-only retrieval across all tested combinations, suggesting robustness to various modality types and numbers. 2) \textit{The Forecasting Contribution Weights} (${\boldsymbol{W}_{\text{TS}}, \cdots, \boldsymbol{W}_{\text{Table}}}$) partially reflect the predictive utility of different modalities: when individually added to TS-only retrieval, text, table, and image receive progressively lower weights, consistent with their decreasing MSE reductions.

More analyses on the sensitivity of the retrieval mechanism for exogenous multimodal information are provided in Appendix~\ref{app: retrieval analysis}.

\section{Conclusion}
In this work, we present QiYao-M, a role-aware multimodal TSFM that distinguishes endo- and exo- modalities according to their forecasting roles. QiYao-M explicitly models endogenous temporal evolution and uses historical-case retrieval to accommodate diverse exo-modality types and numbers without updating TSFM parameters. Experiments on five benchmarks demonstrate strong cross-domain generalization in scenarios both with and without exo-modalities. 
\newpage







\bibliography{bibliography}
\bibliographystyle{iclr2027conference}

\clearpage
\appendix
\section{Experimental Details}
\label{app: exp details}
\subsection{Pretraining Corpus}
\label{app: pretraining corpus}
We construct the pretraining corpus of \name{} from two complementary sources:
a large-scale unimodal time series corpus and a multimodal corpus tailored to
the three-stage training strategy.
The unimodal corpus consists of GIFT-Eval Pretrain, the GIFT-Eval training set~\citep{aksu2024gift},
and the Chronos corpus, providing broad temporal coverage across heterogeneous
domains. Building on this corpus, we further construct 5 million patch-level endogenous
multimodal samples and 10 million sequence-level samples for proxy exogenous
training, supporting the multimodal objectives at different temporal granularities.



\paragraph{Unimodal Pre-training Data.}
The unimodal time series corpus combines three large-scale public data sources:
\begin{itemize}[leftmargin=20pt]
\item \textbf{GIFT-Eval Pretrain.}
We include the pre-training collection released with GIFT-Eval, which contains a diverse set of time series collected from multiple application domains and sampling frequencies. This corpus serves as one of the primary sources of real-world temporal patterns during pre-training.
\item \textbf{GIFT-Eval Training Set.}
In addition to the dedicated pre-training split, we incorporate the training portions of the datasets included in GIFT-Eval~\citep{aksu2024gift}. Only training data are used for pre-training, while the corresponding evaluation portions are kept separated according to the benchmark protocol to avoid test-set leakage.
\item \textbf{Chronos Corpus.}
We further use the large-scale time-series corpus released for Chronos~\citep{ansari2024chronos}, which aggregates time series from heterogeneous real-world sources. The corresponding evaluation portions are kept separated according to the benchmark protocol to avoid test-set leakage.

\end{itemize}

\paragraph{Multimodal Pre-training Data.}
Building on the unimodal time-series corpus, we construct additional multimodal training samples. In total, the multimodal corpus consists of approximately \textbf{5 million patch-level endogenous multimodal samples} and \textbf{10 million sequence-level multimodal samples} used as proxies for exogenous multimodal observations.

\begin{itemize}[leftmargin=20pt]
\item \textbf{Patch-level Endogenous Multimodal Data ($\sim$5M samples).}
We construct multimodal observations at the local patch level, where auxiliary textual and visual information is derived from and aligned with localized temporal patterns within a time series. These samples are designed to establish fine-grained correspondence between numerical dynamics and their multimodal representations. Because the auxiliary modalities describe information intrinsic to the observed time series itself, we refer to them as \emph{endogenous multimodal} data. This corpus is primarily used to strengthen cross-modal representation learning and to encourage the model to recover temporal structures from complementary representations of the same underlying signal.
\item \textbf{Sequence-level Multimodal Data ($\sim$10M samples).}
We additionally construct a substantially larger collection of multimodal samples at the sequence level. Instead of describing individual local patches, the associated textual and visual information is paired with the time series at a coarser sequence-level granularity, providing contextual signals that resemble the external information available in downstream multimodal forecasting scenarios. We therefore use these samples as \emph{proxies for exogenous multimodal training}. This design exposes the model to heterogeneous context during pre-training and encourages it to integrate temporal observations with complementary information beyond local numerical patterns.
\end{itemize}

\subsection{Multimodal Dataset Generation Method}
\label{multimodal_gen}

We construct endogenous text and image modalities directly from numerical
observations using deterministic templates. The resulting modalities therefore provide
complementary views of the same numerical evidence. We generate representations at two granularities: (i)
\emph{patch-level modalities}, which describe fine-grained local dynamics in
each patch, and (ii) \emph{sequence-level modalities}, which summarize
the global evolution of a complete input window.

\paragraph{Mask-aware preprocessing.}
Let a numerical input be
$\mathbf{x}=(x_1,\ldots,x_L)$, with an observation mask
$\mathbf{m}^{\mathrm{obs}}$ and a padding mask
$\mathbf{m}^{\mathrm{pad}}$. We define
\begin{equation}
v_t=m^{\mathrm{obs}}_t(1-m^{\mathrm{pad}}_t),
\end{equation}
where $v_t=1$ indicates an actually observed value. Left padding and naturally
missing observations are recorded separately: padding describes positions
outside the available context, whereas missingness describes unobserved
positions inside the context. Neither is interpreted as an observed zero.
For a patch size $P=32$, the input is divided chronologically into
$K=\lceil L/P\rceil$ non-overlapping patches. If the first patch is incomplete,
it is left-padded, and its padded positions remain masked throughout text and
image generation. During training, historical modalities are generated from historical values
and used as model inputs. Modalities generated from ground-truth future patches
may only be used as auxiliary prediction targets for the \textit{Endo-Multimodal Predictor}. At inference time, only modalities derived from the available
history are constructed; future values or future-generated modalities are
never supplied to the model.

\subsubsection{Patch-Level Endogenous Modalities}

\paragraph{Patch-level text generation.}
For patch $k$, let
$\mathcal{V}_k=\{r:v_{k,r}=1\}$ denote its valid positions. We compute a fixed
set of mask-aware statistics:
\begin{align}
\rho_k & = \frac{|\mathcal{V}_k|}{P},
& \text{(valid-data coverage)}\\
\Delta_k & = x_{k,r_{\mathrm{last}}}-x_{k,r_{\mathrm{first}}},
& \text{(net change)}\\
\beta_k & =
\frac{\sum_{r\in\mathcal{V}_k}(r-\bar r)(x_{k,r}-\bar x_k)}
{\sum_{r\in\mathcal{V}_k}(r-\bar r)^2},
& \text{(least-squares trend)}\\
R_k & = Q_{0.95}(\mathbf{x}_k)-Q_{0.05}(\mathbf{x}_k),
& \text{(robust range)}\\
V_k & = \operatorname{median}_{r}\left|d^{(1)}_{k,r}
-\operatorname{median}(\mathbf{d}^{(1)}_k)\right|,
& \text{(local volatility)}
\end{align}
where quantiles are computed only over valid values. 

The normalized slope, robust range, volatility, lag-one dependence, number of
turning points, and largest standardized jump are mapped to predefined
linguistic bins. These bins determine the fields \texttt{[trend]},
\texttt{[shape]}, \texttt{[variation]}, \texttt{[volatility]},
\texttt{[event]}, and \texttt{[regime]}. The coverage and the two masks
determine \texttt{[padding]}, \texttt{[missingness]}, and
\texttt{[coverage]}. All thresholds are fixed before downstream evaluation;
when data-dependent thresholds are needed, they are estimated from the
training split only.

The categorical fields are inserted into the following deterministic template:
\begin{quote}
\emph{``This patch contains [padding] and [missingness], leaving [coverage]
valid data coverage. Within the valid observations, the series shows [trend]
and follows [shape]. Its local variation is [variation], with [volatility]
volatility. The patch contains [event]. Overall, its local behavior is
characterized as [regime].''}
\end{quote}
This constrained vocabulary makes every statement traceable to a numerical
statistic and prevents unsupported domain-specific interpretations.

Figure~\ref{fig:patch_text_generation_example} shows the intended presentation
of a patch-level example. The middle line explicitly displays the numerical
patch from which the text is generated.

\begin{figure*}[t]
\centering
\begin{generationcase}
\templatefield{This patch contains \texttt{[padding]} and
\texttt{[missingness]}, leaving \texttt{[coverage]} valid data coverage.
Within the valid observations, the series shows \texttt{[trend]} and follows
\texttt{[shape]}. Its local variation is \texttt{[variation]}, with
\texttt{[volatility]} volatility. The patch contains \texttt{[event]}.
Overall, its local behavior is characterized as \texttt{[regime]}.}

\seriesfield{\texttt{[76, 41, 27, 26, 15, 6, 6, 4, 11, 8, 11, 13,
4, 8, 12, 8, 10, 16, 26, 17, 18, 24, 23, 21, 22, 27, 29, 17,
25, 21, 23, 23]}. This is the first 32-point patch of a real
Taobao-Fashion series. It contains no padding or missing observations.}

\generatedfield{This patch contains no padding and no missing observations,
leaving full valid-data coverage. The series starts at 76, drops rapidly to 4
within the first eight positions, and then fluctuates at a lower level before
partially recovering. Its local variation and volatility are high. The patch
contains a prominent early decline and several subsequent turning points.
Overall, its local behavior is characterized as an irregular downward regime.}
\end{generationcase}
\caption{Deterministic patch-level text construction for a real 32-point patch
from Taobao-Fashion item \texttt{948150}. The template, numerical patch, and
generated description are shown together to make the supervision auditable.}
\label{fig:patch_text_generation_example}
\end{figure*}

\paragraph{Patch-level three-channel image generation.}
First- and second-order
differences are defined as
\begin{align}
d^{(1)}_{k,r} &= x_{k,r}-x_{k,r-1},\\
d^{(2)}_{k,r} &= d^{(1)}_{k,r}-d^{(1)}_{k,r-1}
=x_{k,r}-2x_{k,r-1}+x_{k,r-2}.
\end{align}
The corresponding difference is valid only when all values required by the
operation are valid. Hence, differencing never crosses a missing or padded
position.
Each numerical patch is also rendered as a $224\times224$ RGB image. The three
channels encode the value, velocity, and acceleration-like views of the same
patch:
\begin{equation}
\mathcal{I}_k
=\operatorname{Stack}_{\mathrm{RGB}}
\left(\mathcal{R}(\widetilde{\mathbf{x}}_k),
      \mathcal{R}(\widetilde{\mathbf{d}}^{(1)}_k),
      \mathcal{R}(\widetilde{\mathbf{d}}^{(2)}_k)\right),
\end{equation}
where $\mathcal{R}(\cdot)$ denotes mask-aware curve rasterization. The original
value curve is placed in the red channel, the first-order difference curve in
the green channel, and the second-order difference curve in the blue channel.
Their roles are complementary:
\begin{itemize}
    \item \textbf{Value channel (red):} preserves the local level, direction,
    turning points, and overall shape of the observed patch.
    \item \textbf{First-difference channel (green):} represents local increments
    $d^{(1)}_{k,r}$ and exposes the direction and magnitude of short-term
    changes. A nearly horizontal trace indicates approximately constant local
    increments.
    \item \textbf{Second-difference channel (blue):} represents changes in the
    first differences. It highlights curvature, acceleration, deceleration,
    and abrupt changes that may be visually subtle in the raw-value curve.
\end{itemize}

To reduce sensitivity to isolated outliers, the value channel is robustly
normalized using valid patch quantiles:
\begin{equation}
\widetilde{x}_{k,r}
=\operatorname{clip}\left(
\frac{x_{k,r}-Q_{0.50}(\mathbf{x}_k)}
{Q_{0.95}(\mathbf{x}_k)-Q_{0.05}(\mathbf{x}_k)+\epsilon},
-1,1\right).
\end{equation}
The two difference channels share a common robust scale
\begin{equation}
s_{\Delta,k}=Q_{0.95}\left(
\left\{|d^{(1)}_{k,r}|\right\}
\cup
\left\{|d^{(2)}_{k,r}|\right\}
\right)+\epsilon,
\end{equation}
and are normalized by
$\widetilde d^{(j)}_{k,r}=\operatorname{clip}
(d^{(j)}_{k,r}/s_{\Delta,k},-1,1)$. Sharing the scale preserves the relative
magnitudes of the first- and second-order changes. Horizontal coordinates are
determined by the original patch positions, and normalized amplitudes determine
vertical coordinates. Curves are drawn only between consecutive valid
positions. Missing or padded positions remain white and interrupt the curve,
rather than being filled or connected across.

\begin{figure*}[t]
\centering
\begin{generationcase}
\seriesfield{\texttt{[76, 41, 27, 26, 15, 6, 6, 4, 11, 8, 11, 13,
4, 8, 12, 8, 10, 16, 26, 17, 18, 24, 23, 21, 22, 27, 29, 17,
25, 21, 23, 23]}. The red channel encodes the normalized value curve,
the green channel encodes $d^{(1)}_t$, and the blue channel encodes
$d^{(2)}_t$.}

\centering
\includegraphics[width=0.38\linewidth]{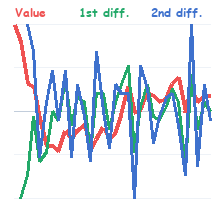}
\end{generationcase}
\caption{Actual $224\times224$ patch-level RGB image generated from the listed
real 32-point Taobao-Fashion patch. The three colored curves encode the value,
first-difference, and second-difference views, respectively.}
\label{fig:patch_image_generation_example}
\end{figure*}

\subsubsection{Sequence-Level Endogenous Modalities}

Patch-level modalities emphasize local behavior but do not explicitly capture
how local regimes evolve over a long context. We therefore additionally
construct sequence-level views from the complete input window. These views use
the same masks, robust statistics, and deterministic vocabulary as their
patch-level counterparts. Unlike patch-level modalities, sequence-level
modalities are not directly fused into the numerical backbone as endogenous
inputs. Instead, they serve as retrieval proxies during the training of the
exo-multimodal retrieval enhancer. Specifically, the generated sequence-level
text and images simulate the heterogeneous exogenous information associated
with a historical time series, allowing the retrieval module to learn how
different exogenous modalities should identify and aggregate relevant
reference cases. This proxy-based training provides scalable supervision for
the exogenous module without requiring every pretraining sequence to be paired
with naturally occurring exogenous multimodal data.

\paragraph{Sequence-level text generation.}
For a complete input window, we compute global coverage, robust range, global
slope, volatility, dominant periodicity, and salient jumps. We also arrange the
patch-level regimes chronologically and compare adjacent patches. A regime
transition is recorded when the trend category, variation category, or turning
behavior changes between neighboring valid patches. These statistics populate
the sequence template:
\begin{quote}
\emph{``This sequence contains [number] patches with [coverage] valid-data
coverage. Across the observed history, it shows [global trend] with [global
variation] variation and [global volatility] volatility. The local dynamics
progress from [early regime] to [late regime], with [transition] near [location].
The sequence contains [global event] and exhibits [periodicity]. Overall, its
temporal evolution is characterized as [global pattern].''}
\end{quote}

\begin{figure*}[t]
\centering
\begin{generationcase}
\templatefield{This sequence contains \texttt{[number]} patches with
\texttt{[coverage]} valid-data coverage. Across the observed history, it shows
\texttt{[global trend]} with \texttt{[global variation]} variation and
\texttt{[global volatility]} volatility. The local dynamics progress from
\texttt{[early regime]} to \texttt{[late regime]}, with \texttt{[transition]}
near \texttt{[location]}. The sequence contains \texttt{[global event]} and
exhibits \texttt{[periodicity]}. Overall, its temporal evolution is
characterized as \texttt{[global pattern]}.}

\seriesfield{{\scriptsize\texttt{[76, 41, 27, 26, 15, 6, 6, 4, 11, 8,
11, 13, 4, 8, 12, 8, 10, 16, 26, 17, 18, 24, 23, 21, 22, 27, 29,
17, 25, 21, 23, 23, 18, 19, 32, 25, 31, 34, 33, 45, 54, 61, 80,
57, 63, 56, 44, 51, 37, 55, 27, 31, 36, 42, 48, 41, 51, 47, 57,
60, 66, 65, 57, 44, 65, 78, 65, 45, 68, 49, 39, 47, 64, 41, 56,
43, 53, 50, 37, 51, 41, 39, 45, 65, 52, 49, 71, 69, 46, 41, 48,
66, 60, 62, 49, 172, 57, 54, 41, 59, 54, 60, 59, 43, 49, 58, 36,
80, 64, 58, 36, 40, 33, 47, 73, 58, 85, 54, 41, 52, 52, 62, 78,
61, 51, 60, 109, 43, 48, 74, 35, 37, 51, 48, 41, 47, 46, 51, 22,
34, 25, 44, 37, 37, 22, 18, 18, 33, 35, 40, 39, 27, 41, 39, 30,
30, 26, 35, 26, 33, 25, 46, 29, 39, 40, 44, 31, 33, 38, 54, 41,
51, 49, 50, 39, 38, 36, 57, 37, 34, 47, 34, 54, 54, 47, 37, 16,
15, 6, 5, 5, 3, 2, 1, 4, 7, 12, 11, 14, 23, 30, 53, 69, 76, 85,
92, 95, 75, 61, 73, 81, 83, 155, 150, 129, 149, 133, 30, 1, 1, 1,
0, 0, 0, 0, 1, 0, 0, 0, 0, 0, 0, 0, 10, 11, 9, 9, 10, 5, 7,
10, 8, 8, 9, 8, 4, 13, 8, 6, 6, 11, 10, 7, 13, 9, 7]}}.
This is a real 256-day window from Taobao-Fashion item \texttt{948150},
covering 2014-08-08 to 2015-04-20. It contains eight fully observed
32-point patches; zero-valued observations are retained as valid values.}

\generatedfield{This sequence contains eight patches with full valid-data
coverage. Across the observed history, it shows substantial variation and
high volatility, with several distinct local regimes. It begins with a sharp
decline, moves into a sustained medium-level fluctuating regime, and contains
a prominent spike of 172 near the end of the third patch. The middle patches
gradually return to lower values. A later rebound reaches 155, followed by an
abrupt collapse into a near-zero regime and a small recovery near the end.
The repeated changes in level dominate any stable periodic pattern. Overall,
the sequence is characterized as a non-stationary multi-regime trajectory
with multiple abrupt transitions.}
\end{generationcase}
\caption{Deterministic sequence-level text construction from a real 256-point
Taobao-Fashion input window. The item identifier and date range are retained
to make the example traceable to the source data.}
\label{fig:sequence_text_generation_example}
\end{figure*}

\paragraph{Sequence-level line image.}
For the complete sequence, we draw the valid value, first-difference, and
second-difference curves in chronological order. The construction follows the
same definitions as the patch-level RGB image, but uses the full historical
window and a sequence-level robust scale. This view makes long-term trend,
regime changes, repeated oscillations, and change points visible in a single
image. Missing intervals remain blank, and curves are not connected across
them.

\paragraph{Sequence-level heatmap.}
To expose relationships across patches, the sequence is reshaped into a
$K\times P$ patch matrix. For valid position $(k,r)$, we set
\begin{equation}
H_{k,r}=\operatorname{clip}\left(
\frac{x_{k,r}-Q_{0.50}(\mathbf{x})}
{Q_{0.75}(\mathbf{x})-Q_{0.25}(\mathbf{x})+\epsilon},-3,3\right).
\end{equation}
Rows correspond to chronological patches and columns correspond to within-patch
positions. A fixed diverging color map represents negative and positive
deviations from the sequence median; masked entries are rendered white. Because
the normalization is shared across the whole sequence rather than performed
independently for every row, differences in level and amplitude remain
comparable across patches.

\paragraph{Sequence-level spectrum image.}
The frequency-domain view describes periodic and multi-scale behavior that may
be difficult to identify in the time-domain curves. For a contiguous valid
block $\mathcal{B}$, we remove its robust center, apply a Hann window $w_t$, and
compute the periodogram
\begin{equation}
S_{\mathcal{B}}(f)=
\frac{\left|\sum_{t\in\mathcal{B}}w_t
(x_t-\bar{x}_{\mathcal{B}})e^{-\mathrm{i}2\pi ft}\right|^2}
{\sum_{t\in\mathcal{B}}w_t^2+\epsilon}.
\end{equation}
When missing intervals exist, periodograms are computed separately for valid
contiguous blocks and combined on a common normalized-frequency grid, weighted
by block length. We never interpolate across a missing interval solely to make
the Fourier transform visually continuous. The plotted vertical coordinate is
$\log(1+S(f))$, which compresses extreme spectral peaks while retaining the
dominant frequencies.

\begin{figure*}[t]
\centering
\begin{subfigure}[t]{0.32\textwidth}
\centering
\includegraphics[width=\linewidth]{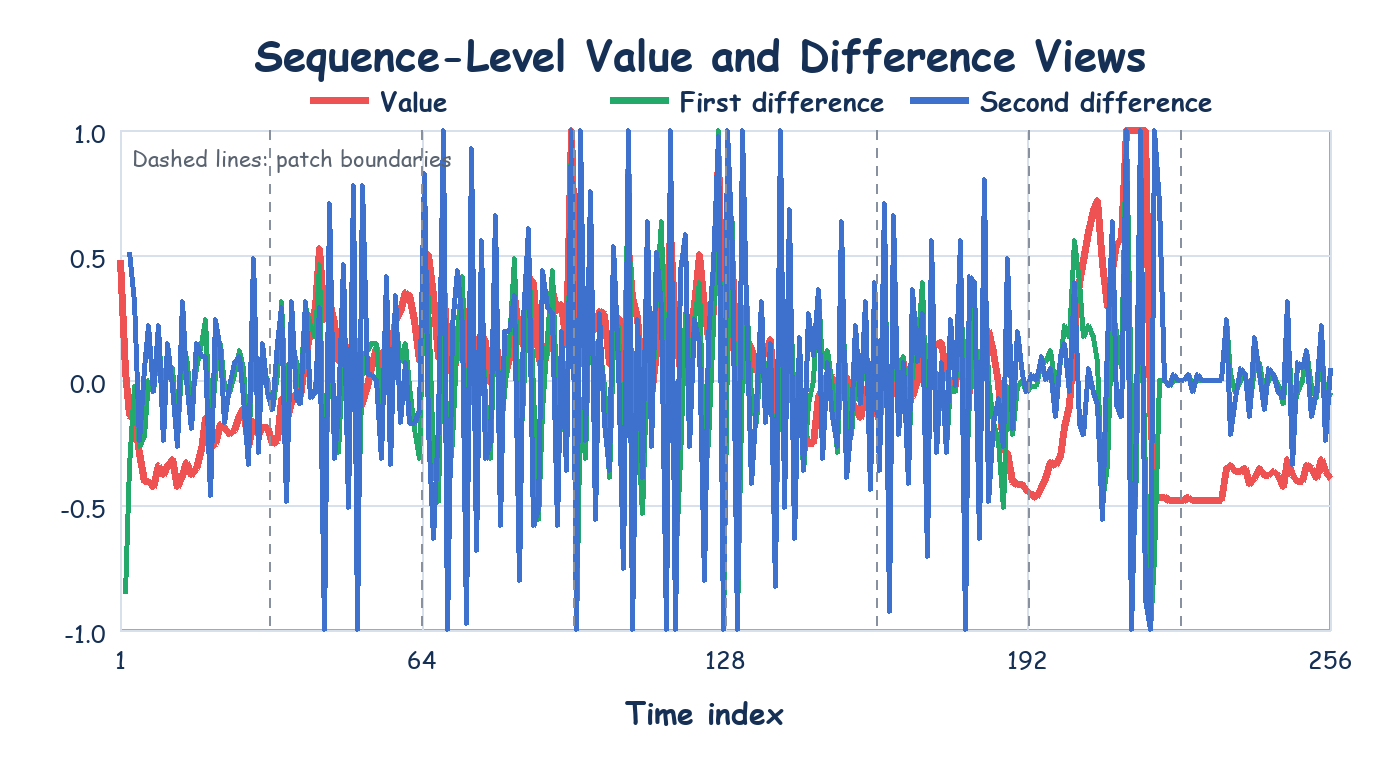}
\caption{Value, first-difference, and second-difference curves.}
\end{subfigure}\hfill
\begin{subfigure}[t]{0.32\textwidth}
\centering
\includegraphics[width=\linewidth]{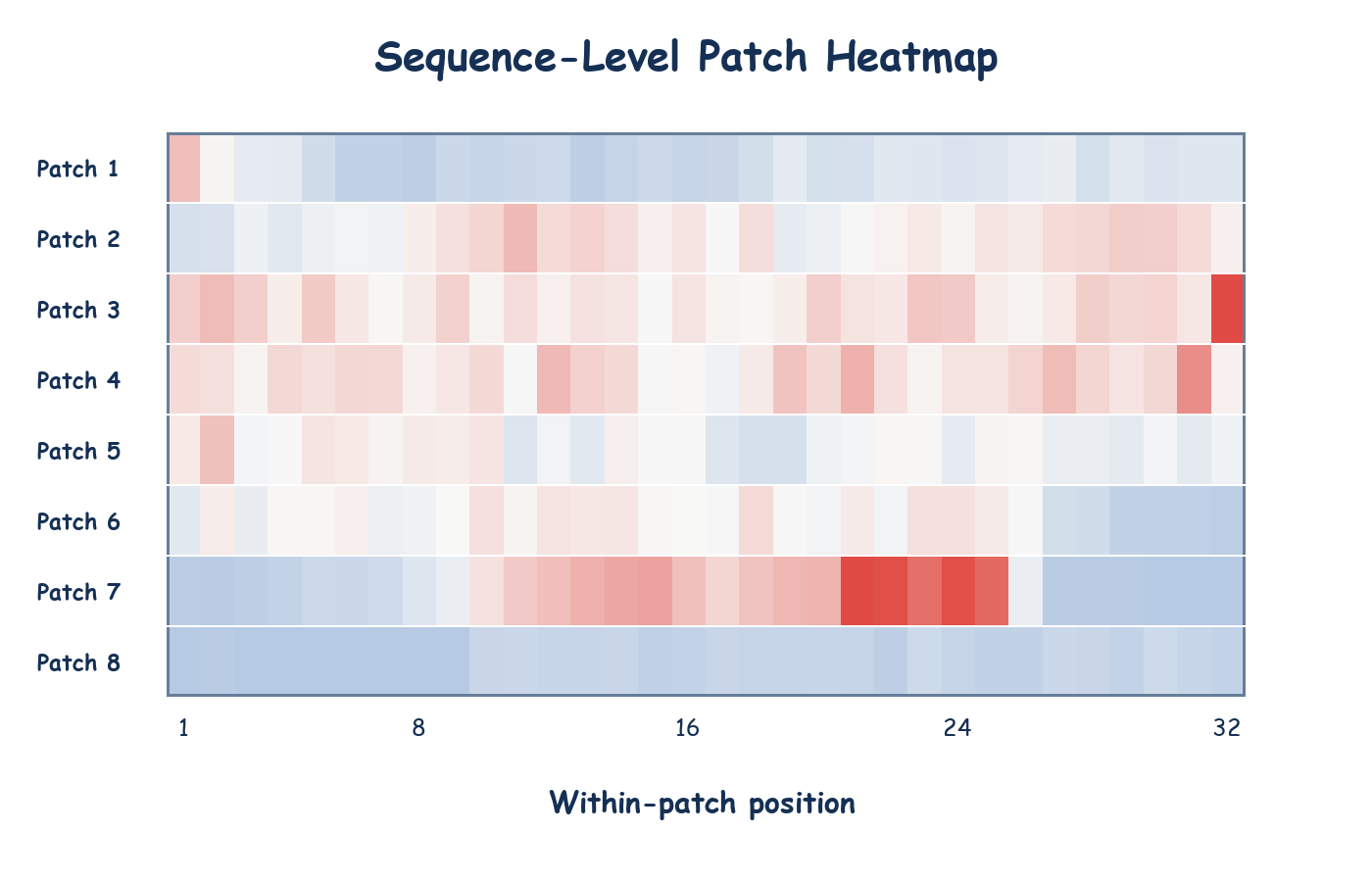}
\caption{Mask-aware patch-by-position heatmap.}
\end{subfigure}\hfill
\begin{subfigure}[t]{0.32\textwidth}
\centering
\includegraphics[width=\linewidth]{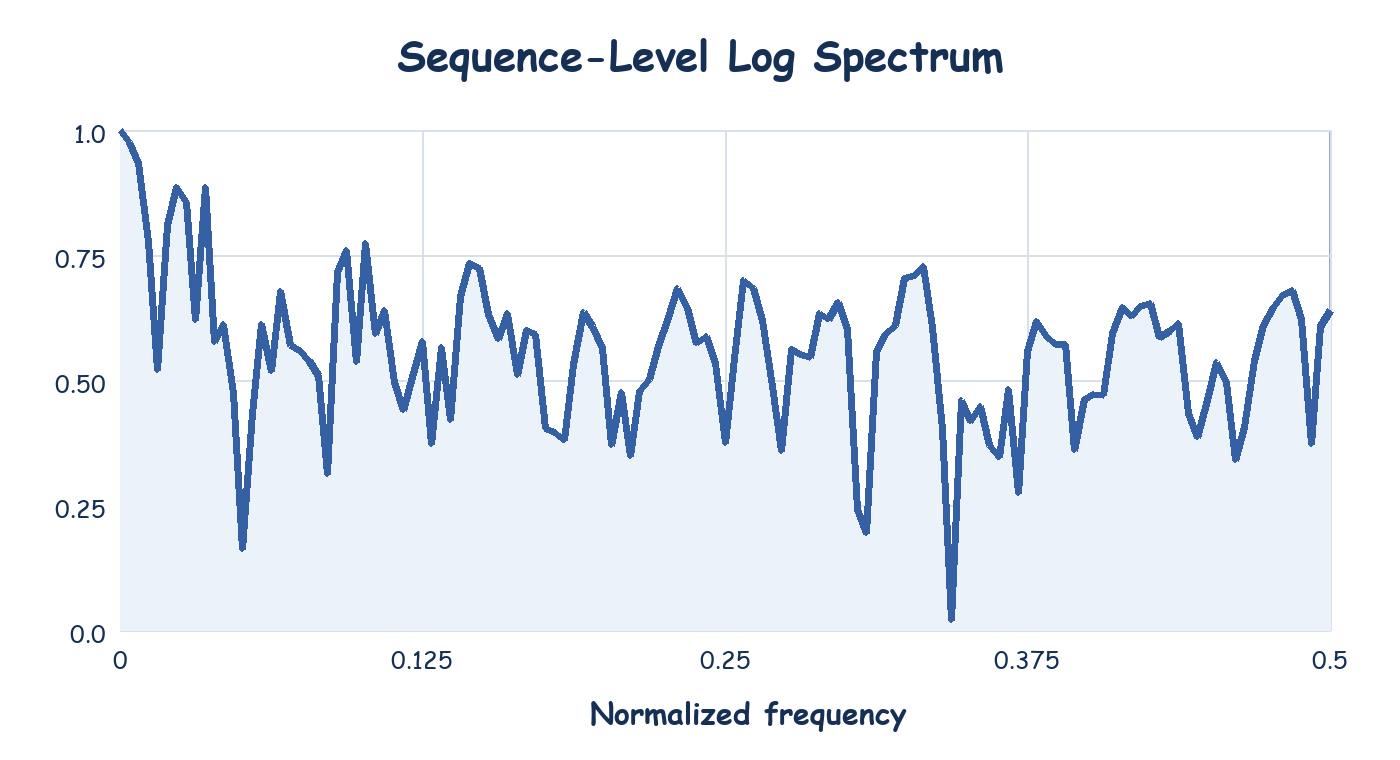}
\caption{Log-scaled frequency spectrum.}
\end{subfigure}
\caption{The three sequence-level image views generated deterministically from
the same real 256-point Taobao-Fashion historical sequence.}
\label{fig:sequence_image_generation_examples}
\end{figure*}

\paragraph{Dataset construction and reproducibility.}
For every training window, we store the numerical values, observation mask,
padding mask, patch boundaries, patch-level text, patch-level RGB images,
sequence-level text, and the three sequence-level image views. Each artifact is
associated with the source identifier, window endpoint, and generation-version
identifier. Identical numerical inputs and masks therefore produce identical
multimodal examples. Dataset splitting is performed chronologically before
generation statistics are estimated, and any threshold or normalization
summary shared across examples is obtained from the training partition only.
This construction preserves the forecasting information boundary while
providing aligned local and global multimodal supervision.

\subsection{Benchmarks}
\label{app: benchmarks}
We evaluate \name{} on both unimodal and multimodal forecasting benchmarks. For unimodal forecasting, we adopt GIFT-Eval~\citep{aksu2024gift} and TIME~\citep{qiao2026s}. For forecasting with exogenous modalities, we consider representative multimodal benchmarks covering increasingly diverse modality combinations: Time-MMD~\citep{liu2024time} (time series + text), MoTime~\citep{zhou2025motime} (time series + text + image), and FinMultiTime~\citep{xu2025finmultitime} (time series + text + image + table).


\paragraph{Endogenous Multimodal Forecasting.}
We evaluate forecasting without externally provided modalities on
GIFT-Eval and TIME.
In this setting, multimodal representations are constructed solely from the observed
time series, allowing us to examine whether endogenous multimodal modeling improves
forecasting even in the absence of external information.
GIFT-Eval contains 23 datasets spanning diverse domains and sampling frequencies,
while TIME consists of 50 datasets, resulting in a total of
\textbf{73 evaluation datasets}.
We follow the official evaluation protocols of the two benchmarks, including their
predefined forecasting horizons, sampling frequencies, and evaluation windows.

\paragraph{Exogenous Multimodal Forecasting.}
To evaluate the ability of \name{} to exploit externally provided information,
we further consider 9 multimodal datasets from Time-MMD, MoTime,
and FinMultiTime.
Specifically, we use
Agriculture, Climate, Energy,
Health, and Social Good from Time-MMD;
Taobao-Fashion and Tianchi from MoTime;
and HS300 and SP500 from FinMultiTime.
These datasets cover different combinations of time series, text, images,
and tabular information, enabling evaluation under heterogeneous
multimodal settings. For Agriculture, Climate, and Social Good, we use prediction lengths
$F \in \{6, 8, 10, 12\}$.
For Energy and Health, the prediction lengths are
$F \in \{12, 24, 36, 48\}$.
For Taobao-Fashion and Tianchi, we evaluate
$F \in \{1, 7, 14, 21, 28\}$.
For HS300 and SP500, the prediction lengths are
$F \in \{24, 48, 96\}$.

\begin{table*}[h]
\centering
\caption{
Dataset statistics and forecasting configurations of GIFT-Eval.
Multiple frequencies of the same underlying dataset are grouped into one row.
}
\label{tab:gifteval_datasets}

\scriptsize
\setlength{\tabcolsep}{4.2pt}
\renewcommand{\arraystretch}{1.08}

\resizebox{\textwidth}{!}{
\begin{tabular}{l l l c l}
\toprule

\textbf{Dataset}
& \textbf{Domain}
& \textbf{Frequency}
& \textbf{Variates}
& \textbf{Prediction Length} \\

\midrule

Jena Weather
& Nature
& 10T / H / D
& 21
& $\{48,480,720\}$ (10T/H); 30 (D) \\

BizITObs--Application
& Web/CloudOps
& 10S
& 2
& $\{60,600,900\}$ \\

BizITObs--Service
& Web/CloudOps
& 10S
& 2
& $\{60,600,900\}$ \\

BizITObs--L2C
& Web/CloudOps
& 5T / H
& 7
& $\{48,480,720\}$ \\

Bitbrains--Fast Storage
& Web/CloudOps
& 5T / H
& 2
& $\{48,480,720\}$ (5T); 48 (H) \\

Bitbrains--rnd
& Web/CloudOps
& 5T / H
& 2
& $\{48,480,720\}$ (5T); 48 (H) \\

Restaurant
& Sales
& D
& 1
& 30 \\

ETT1
& Energy
& 15T / H / D / W
& 7
& $\{48,480,720\}$ (15T/H); 30 (D); 8 (W) \\

ETT2
& Energy
& 15T / H / D / W
& 7
& $\{48,480,720\}$ (15T/H); 30 (D); 8 (W) \\

Loop Seattle
& Transport
& 5T / H / D
& 1
& $\{48,480,720\}$ (5T/H); 30 (D) \\

SZ-Taxi
& Transport
& 15T / H
& 1
& $\{48,480,720\}$ (15T); 48 (H) \\

M\_DENSE
& Transport
& H / D
& 1
& $\{48,480,720\}$ (H); 30 (D) \\

Solar
& Energy
& 10T / H / D / W
& 1
& $\{48,480,720\}$ (10T/H); 30 (D); 8 (W) \\

Hierarchical Sales
& Sales
& D / W
& 1
& 30 (D); 8 (W) \\

M4
& Econ/Fin
& A / Q / M / W / D / H
& 1
& 6 / 8 / 18 / 13 / 14 / 48 \\

Hospital
& Healthcare
& M
& 1
& 12 \\

COVID Deaths
& Healthcare
& D
& 1
& 30 \\

US Births
& Healthcare
& D / W / M
& 1
& 30 / 8 / 12 \\

Saugeen
& Nature
& D / W / M
& 1
& 30 / 8 / 12 \\

Temperature Rain
& Nature
& D
& 1
& 30 \\

KDD Cup 2018
& Nature
& H / D
& 1
& $\{48,480,720\}$ (H); 30 (D) \\

Car Parts
& Sales
& M
& 1
& 12 \\

Electricity
& Energy
& 15T / H / D / W
& 1
& $\{48,480,720\}$ (15T/H); 30 (D); 8 (W) \\

\bottomrule
\end{tabular}
}

\end{table*}

\begin{table*}[h]
\centering

\caption{
Dataset statistics and forecasting configurations of the TIME benchmark.
Different sampling frequencies of the same underlying source are treated as
separate evaluation datasets.
}
\label{tab:time_datasets}

\scriptsize
\setlength{\tabcolsep}{4.5pt}
\renewcommand{\arraystretch}{0.90}

\begin{adjustbox}{
    max width=\textwidth,
    max totalheight=0.88\textheight,
    keepaspectratio,
    center
}

\begin{tabular}{
    >{\centering\arraybackslash}p{0.14\textwidth}
    >{\raggedright\arraybackslash}p{0.30\textwidth}
    >{\centering\arraybackslash}p{0.12\textwidth}
    >{\centering\arraybackslash}p{0.24\textwidth}
}

\toprule

\textbf{Domain}
& \textbf{Dataset}
& \textbf{Frequency}
& \textbf{Prediction Length} \\

\midrule


\multirow{16}{*}{Nature}
& Water Quality Darwin
& 15T
& $\{16, 96, 288\}$ \\

& Current Velocity
& 5T
& $\{36, 288, 864\}$ \\

& Current Velocity
& 10T
& $\{18, 144, 432\}$ \\

& Current Velocity
& 15T
& $\{12, 96, 288\}$ \\

& Current Velocity
& 20T
& $\{9, 72, 216\}$ \\

& Current Velocity
& H
& $\{24, 168, 336\}$ \\

& CPHL
& 15T
& $\{12, 96, 288\}$ \\

& CPHL
& 30T
& $\{12, 48, 144\}$ \\

& CPHL
& H
& $\{24, 168, 336\}$ \\

& Coastal T\&S
& 5T
& $\{36, 288, 864\}$ \\

& Coastal T\&S
& 15T
& $\{12, 96, 288\}$ \\

& Coastal T\&S
& 20T
& $\{9, 72, 216\}$ \\

& Coastal T\&S
& H
& $\{24, 168, 336\}$ \\

& SG Weather
& D
& $\{3, 7, 14\}$ \\

& SG PM2.5
& H
& $\{24, 72, 168\}$ \\

& NE China Wind
& H
& $\{24, 72, 168\}$ \\

\midrule


\multirow{4}{*}{Energy}
& Australia Solar
& H
& $\{24, 72, 168\}$ \\

& EPF Electricity Price
& H
& $\{24, 72, 168\}$ \\

& OpenElectricity NEM
& 5T
& $\{24, 96, 288\}$ \\

& EWELD Load
& 15T
& $\{24, 96, 672\}$ \\

\midrule


\multirow{4}{*}{Transport}
& SG Carpark
& 15T
& $\{16, 96, 672\}$ \\

& Finland Traffic
& 15T
& $\{16, 96, 672\}$ \\

& Port Activity
& D
& 30 \\

& Port Activity
& W
& 13 \\

\midrule


\multirow{3}{*}{Healthcare}
& ECDC COVID
& D
& 30 \\

& ECDC COVID
& W
& 13 \\

& Global Influenza
& W
& 13 \\

\midrule


\multirow{3}{*}{Finance}
& Crypto
& D
& 30 \\

& US Term Structure
& B
& 20 \\

& Oil Price
& B
& 20 \\

\midrule


\multirow{11}{*}{Economics}
& Job Claims
& W
& 13 \\

& Uncertainty 1M
& M
& 6 \\

& Housing Inventory
& M
& 12 \\

& JOLTS
& M
& 12 \\

& US Labor
& M
& 12 \\

& Vehicle Supply
& M
& 12 \\

& Auto Production SF
& M
& 12 \\

& Commodity Production
& M
& 12 \\

& Commodity Import
& M
& 12 \\

& WUI Global
& Q
& 4 \\

& Global Price
& Q
& 4 \\

\midrule


\multirow{4}{*}{Sales}
& Vehicle Sales
& M
& 12 \\

& Online Retail II UCI
& D
& 30 \\

& Supply Chain Customer
& D
& 30 \\

& Supply Chain Location
& D
& 30 \\

\midrule


\multirow{5}{*}{CloudOps}
& Azure2019 D
& 5T
& 288 \\

& Azure2019 I
& 5T
& 288 \\

& Azure2019 U
& 5T
& 48 \\

& Smart Manufacturing
& H
& $\{24, 168, 336\}$ \\

& MetroPT-3
& 5T
& $\{48, 288, 576\}$ \\

\bottomrule

\end{tabular}

\end{adjustbox}

\end{table*}

\begin{table*}[h]
\centering
\caption{
Statistics of the exogenous multimodal forecasting datasets used in our experiments.
The modality column reports the external information available to each dataset.
}
\label{tab:multimodal_datasets}

\small
\setlength{\tabcolsep}{4.5pt}
\renewcommand{\arraystretch}{1.08}

\resizebox{\textwidth}{!}{
\begin{tabular}{l l l l l l}
\toprule

\textbf{Benchmark}
& \textbf{Dataset}
& \textbf{Modalities}
& \textbf{Frequency}
& \textbf{Scale}
& \textbf{Prediction Length} \\

\midrule

\multirow{5}{*}{TimeMMD}
& Agriculture
& TS + Text
& Monthly
& $496 \times 1$
& $\{6,8,10,12\}$ \\

& Climate
& TS + Text
& Monthly
& $496 \times 5$
& $\{6,8,10,12\}$ \\

& Energy
& TS + Text
& Weekly
& $1{,}479 \times 9$
& $\{12,24,36,48\}$ \\

& Health
& TS + Text
& Weekly
& $1{,}389 \times 11$
& $\{12,24,36,48\}$ \\

& Social Good
& TS + Text
& Monthly
& $900 \times 1$
& $\{6,8,10,12\}$ \\

\midrule

\multirow{2}{*}{MoTIME}
& TaobaoFashion
& TS + Text + Image
& Daily
& $365 \times 890$
& $\{1,7,14,21,28\}$ \\

& Tianchi
& TS + Text + Image
& Daily
& $184 \times 36{,}397$
& $\{1,7,14,21,28\}$ \\

\midrule

\multirow{2}{*}{FinMultiTime}
& HS300
& TS + Text + Image + Table
& Trading Day
& 35 stocks
& $\{24,48,96\}$ \\

& SP500
& TS + Text + Image + Table
& Trading Day
& 35 stocks
& $\{24,48,96\}$ \\

\bottomrule
\end{tabular}
}

\end{table*}

\subsection{Baselines}
\label{app: baselines}

We compare \name{} against representative 
forecasting models
from three categories.
For GIFT-Eval and TIME, we use baseline results reported on the official leaderboards
under the corresponding benchmark protocols, which ensures consistent and standardized
evaluation across methods. For multimodal forecasting, we consider four end-to-end supervised models, including
GPT4MTS~\citep{jia2024gpt4mts},
CALF~\citep{liu2025calf},
Time-VLM~\citep{time-vlm},
and TATS~\citep{tats}.
We further include five unimodal time-series foundation models, namely
Zeus~\citep{zeus},
Chronos-2~\citep{ansari2025chronos2},
Toto-2~\citep{toto2},
PatchTST-FM-r2~\citep{patchtstfm},
and TiRex-2~\citep{tirex2},
as well as two multimodal foundation models,
ChatTime~\citep{chattime} and Aurora~\citep{wu2026aurora}.
The codebases and implementation details of all baselines are provided in
Table~\ref{tab:baselines}.
\begin{table}[!htbp]
    \centering
    \caption{
    Code and model repositories of the baseline methods used in our experiments.
    }
    \label{tab:baselines}

    \small
    \setlength{\tabcolsep}{4pt}
    \renewcommand{\arraystretch}{1.15}
    \urlstyle{same}

    \begin{tabularx}{\linewidth}{
        >{\raggedright\arraybackslash}p{0.22\linewidth}
        >{\raggedright\arraybackslash}p{0.18\linewidth}
        >{\raggedright\arraybackslash}X
    }
        \toprule
        \textbf{Model Type}
        & \textbf{Model}
        & \textbf{Code / Model Repository} \\
        \midrule

        \multirow{4}{*}{\shortstack[l]{Multimodal\\End-to-End}}
        & GPT4MTS
        & \href{https://github.com/Flora-jia-jfr/GPT4MTS-Prompt-based-Large-Language-Model-for-Multimodal-Time-series-Forecasting}
        {\shortstack[l]{
        https://github.com/Flora-jia-jfr/\\
        GPT4MTS-Prompt-based-Large-Language-Model-\\
        for-Multimodal-Time-series-Forecasting
        }} \\
        
        \cmidrule(l){2-3}

        & CALF
        & \url{https://github.com/Hank0626/CALF} \\

        \cmidrule(l){2-3}

        & Time-VLM
        & \url{https://github.com/CityMind-Lab/ICML25-TimeVLM} \\

        \cmidrule(l){2-3}

        & TATS
        & \url{https://github.com/iDEA-iSAIL-Lab-UIUC/TaTS} \\

        \midrule

        \multirow{5}{*}{\shortstack[l]{Unimodal\\Foundation}}
        & Zeus
        & \url{https://github.com/GestaltCogTeam/Zeus} \\

        \cmidrule(l){2-3}

        & Chronos-2
        & \url{https://github.com/amazon-science/chronos-forecasting} \\

        \cmidrule(l){2-3}

        & Toto-2.0
        & \url{https://github.com/DataDog/toto} \\

        \cmidrule(l){2-3}

        & PatchTST-FM-r2
        & \url{https://huggingface.co/ibm-granite/granite-timeseries-patchtst-fm-r2} \\

        \cmidrule(l){2-3}

        & TiRex-2
        & \url{https://github.com/NX-AI/tirex-2} \\

        \midrule

        \multirow{2}{*}{\shortstack[l]{Multimodal\\Foundation}}
        & ChatTime
        & \url{https://github.com/ForestsKing/ChatTime} \\

        \cmidrule(l){2-3}

        & Aurora
        & \url{https://github.com/decisionintelligence/Aurora} \\

        \bottomrule
    \end{tabularx}

\end{table}

\subsection{Pretraining Settings}
\label{app: pretraining settings}

\paragraph{Stage I: Numerical Pretraining.}
We first train the time series backbone using numerical data only. Historical
series are normalized and divided into patches of size $p=32$, with a maximum
input length of 8192 observations. The backbone and forecasting head learn
general temporal representations and probabilistic forecasting before any
multimodal components are introduced. This stage provides the numerical
initialization for the subsequent stages.

\paragraph{Stage II: Endogenous Multimodal Training.}
We then construct patch-level text and images from the time series using the
deterministic generation procedure in Section~\ref{multimodal_gen}. The first
16 Transformer layers are frozen, while the remaining 8 layers and the
endogenous multimodal modules are trained. The model jointly predicts future
numerical values and the latent representations of future endogenous text and
images. This design encourages the shared representation to capture temporal
evolution from complementary numerical, textual, and visual perspectives,
while preserving the general forecasting ability learned in Stage I.

\paragraph{Stage III: Exogenous Retrieval Training.}
Finally, we freeze the numerical backbone and the endogenous multimodal
modules. Only the exogenous retrieval head and its candidate-aware
cross-attention components are optimized. Reference cases are retrieved
independently through the available modalities, and their known future
values provide retrieval-based context for the forecast. Retrieval banks and
modality weights are constructed or calibrated using the permitted training
data; test targets are never used for training or calibration. A null
key--value candidate allows the cross-attention module to reduce its reliance
on unhelpful retrieved cases.

\paragraph{Downstream Forecasting.}
At inference time, endogenous text and images are generated solely from the
observed history. For the exogenous setting, reference cases are selected
from the preconstructed bank, and the trained retrieval module enhances the
forecast without updating model parameters. We report metrics over all valid
forecast points and set drop\_last=False during evaluation so that
incomplete final batches are not discarded.

\subsection{Evaluation Metrics}
\label{app:evaluation_metrics}

We follow the official evaluation protocols of the respective benchmarks.
For point forecasting, we report Mean Squared Error (MSE) and Mean Absolute
Error (MAE). For probabilistic forecasting, we report Mean Absolute Scaled
Error (MASE) and Continuous Ranked Probability Score (CRPS) where required.
All metrics are computed over valid forecast points only; padded or missing
targets are excluded. Let $y_{i,t}$ and $\hat y_{i,t}$ denote the target and
point forecast for series $i$ at future step $t$, and let $m_{i,t}\in\{0,1\}$
indicate whether that target is valid. Define
$N=\sum_{i,t}m_{i,t}$.

\paragraph{Point forecasting metrics.}
The MSE penalizes large prediction errors more strongly, whereas the MAE
measures their average absolute magnitude:
\begin{equation}
\mathrm{MSE}
=\frac{1}{N}\sum_{i,t}m_{i,t}(y_{i,t}-\hat y_{i,t})^2,
\qquad
\mathrm{MAE}
=\frac{1}{N}\sum_{i,t}m_{i,t}|y_{i,t}-\hat y_{i,t}|.
\end{equation}
For models producing predictive quantiles, we use the median ($0.5$
quantile) as the point forecast.

\paragraph{Scale-normalized metrics.}
When reporting normalized errors, we calculate the standard deviation
$\sigma_{\mathrm{train}}$ from the training interval only and apply the same
scale to every model evaluated on that dataset:
\begin{equation}
\mathrm{Norm\text{-}MSE}
=\frac{\mathrm{MSE}}{\sigma_{\mathrm{train}}^2},
\qquad
\mathrm{Norm\text{-}MAE}
=\frac{\mathrm{MAE}}{\sigma_{\mathrm{train}}}.
\end{equation}
Normalization changes only the reported metric; it does not use test-set
statistics or alter the model forecasts.

\paragraph{Mean Absolute Scaled Error.}
MASE scales the absolute forecast error by the in-sample seasonal naive
error. For series $i$ with training observations $x_{i,1:L_i}$ and
seasonality $s_i$, the scaling factor is
\begin{equation}
d_i=\frac{1}{L_i-s_i}
\sum_{u=s_i+1}^{L_i}|x_{i,u}-x_{i,u-s_i}|,
\qquad
\mathrm{MASE}
=\frac{1}{N}\sum_{i,t}m_{i,t}
\frac{|y_{i,t}-\hat y_{i,t}|}{d_i}.
\end{equation}
The benchmark's official evaluator determines the seasonality and handles
degenerate denominators.

\paragraph{Continuous Ranked Probability Score.}
For a predictive cumulative distribution function $F_{i,t}$, CRPS compares
the full forecast distribution with the realized target:
\begin{equation}
\mathrm{CRPS}
=\frac{1}{N}\sum_{i,t}m_{i,t}
\int_{-\infty}^{\infty}
\left(F_{i,t}(z)-\mathbb{I}\{y_{i,t}\leq z\}\right)^2\,dz.
\end{equation}
We use each benchmark's official evaluator to compute its reported CRPS
from the model's probabilistic forecasts.

\subsection{Model Configurations}
\label{app:model_configurations}

\begin{table}[!htbp]
\centering
\caption{Architecture and parameter counts of QiYao-M. The main-model count
excludes the frozen CLIP feature encoder and does not double-count the
shared forecasting head.}
\resizebox{\linewidth}{!}{
\begin{tabular}{lccccccccc}
\toprule
\textbf{Model} &
\textbf{Backbone Layers} &
\textbf{Fusion After} &
\textbf{Model Dim.} &
\textbf{FFN Dim.} &
\textbf{Heads} &
\textbf{Patch Size} &
\textbf{CLIP Dim.} &
\textbf{Retrieval Layers} &
\textbf{Main Parameters} \\
\midrule
\name{} & 24 & 16 & 1024 & 2736 & 16 & 32 & 512 & 5 & 375.63M \\
\bottomrule
\end{tabular}}
\label{tab:morrow_config}
\end{table}

The numerical backbone contains 312.68M parameters. The endogenous image
and text projectors, predictors, and mask tokens add 36.73M parameters.
The exogenous target encoder and five candidate-aware cross-attention layers
add 26.21M parameters, yielding 375.63M unique parameters in the main
model. The frozen CLIP ViT-B/32 image
and text encoder contains an additional 151.28M parameters; including it
gives 526.90M parameters.

\section{Related work}

\subsection{Time Series Forecasting}
Time series analysis has witnessed rapid progress across diverse tasks, including
forecasting~\citep{qiu2024tfb,wu2025k2vae,cheng2026kite},
anomaly detection~\citep{wu2025catch,qiu2025tab,cheng2026star},
and emerging reasoning-oriented tasks~\citep{lu2026patra,wu2026timeart}.
Among them, forecasting~\citep{hou2025graph,ding2026timemosaic} has received extensive attention across a variety of
settings and modeling perspectives.
Recent studies improve multivariate forecasting by capturing cross-variable
dependencies and learning effective temporal representations~\citep{cheng2026ccd,qiu2025duet,wu2025srsnet},
while other efforts address irregularly sampled and non-stationary time
series~\citep{liu2026rethinking,qiu2026bridging,lu2026dtaf,liu2026astgi}.
Another important direction incorporates exogenous variables to enrich the
information available for target prediction~\citep{li2026gcgnet,cheng2026kite,qiu2026dag}.
Meanwhile, probabilistic forecasting explicitly characterizes predictive
uncertainty~\citep{wu2025k2vae,cheng2026kite}, and benchmark studies facilitate
systematic evaluation across diverse forecasting scenarios~\citep{zhu2026wpbench,qiu2024tfb}.

\subsection{Multimodal Time Series Forecasting}
Recent studies have increasingly explored multimodal information for time series forecasting. One line of work bridges temporal and language representations by adapting pretrained language models to time series, including GPT4TS~\citep{gpt4ts}, TEST~\citep{test}, Time-LLM~\citep{timellm}, CALF~\citep{liu2025calf}, LLM-Mixer~\citep{llmmixer}, and CC-Time~\citep{cctime}, through reprogramming, representation alignment, or cross-modal interaction. Another line explicitly incorporates auxiliary textual information into forecasting. CMIN~\citep{cmin} and Modality-aware Transformer~\citep{modality-aware} jointly model financial sequences with news or textual reports, while GPT4MTS~\citep{jia2024gpt4mts}, Time-MMD~\citep{time-mmd}, TaTS~\citep{tats}, and VoT~\citep{wang2026vot} further explore the collection, alignment, fusion, or reasoning of contextual text with time series. 
Beyond text, Time-VLM~\citep{time-vlm} introduces visual representations to complement temporal and textual information.  
Despite this progress, most existing approaches learn multimodal forecasting through task- or dataset-specific adaptation, 
rather than large-scale multimodal pretraining across diverse time series domains, limiting the transfer of multimodal knowledge to unseen domains.

\subsection{Time Series Foundation Models}
Time series foundation models (TSFMs) leverage large-scale pretraining to learn transferable temporal patterns and enable zero-shot generalization across domains. Representative models, including  UniTS~\citep{units}, TimesFM~\citep{timesfm}, 
LightGTS~\citep{wang2025lightgts}, PatchTST-FM~\citep{patchtstfm}, Sundial~\citep{liu2025sundial}, 
and ROSE~\citep{wang2025rose}, explore diverse architectures and pretraining objectives for general-purpose forecasting. More recent models further broaden their capabilities: Toto 2.0~\citep{toto2} investigates large-scale model scaling, Chronos-2~\citep{ansari2025chronos2} supports multivariate and covariate-informed forecasting, TiRex-2~\citep{tirex2} enables efficient recurrent and streaming forecasting, and ZEUS~\citep{zeus} extends pretraining toward multiple time series tasks. Recently, ChatTime~\citep{chattime}, STRIDE~\citep{ahamed2026stride} and Aurora~\citep{wu2026aurora} further introduce multimodal pretraining and modeling into TSFMs. However, these models primarily support exogenous textual information and struggle
to generalize to scenarios with various types and numbers of exo-modalities.
More importantly, existing multimodal TSFMs typically adopt role-agnostic modeling,
using  shared mechanisms across heterogeneous modalities without explicitly
modeling endogenous  and exogenous  modalities according to their distinct forecasting characteristics.

In this work, we propose \name{}, a role-aware multimodal TSFM that explicitly distinguishes endogenous and exogenous modalities according to their distinct roles in forecasting. Through role-aware multimodal modeling, \name{} learns how endogenous modalities evolve along with the underlying time series during pretraining. For exogenous modalities, \name{} directly leverages retrieved reference cases as in-context demonstrations at inference time, enabling tuning-free acquisition of domain-specific knowledge and flexible adaptation to arbitrary exogenous modalities.

\clearpage

\section{More Analysis}
\subsection{Retrieval Analysis}
\label{app: retrieval analysis}

\begin{figure*}[h]
  \centering
\includegraphics[width=0.7\columnwidth]{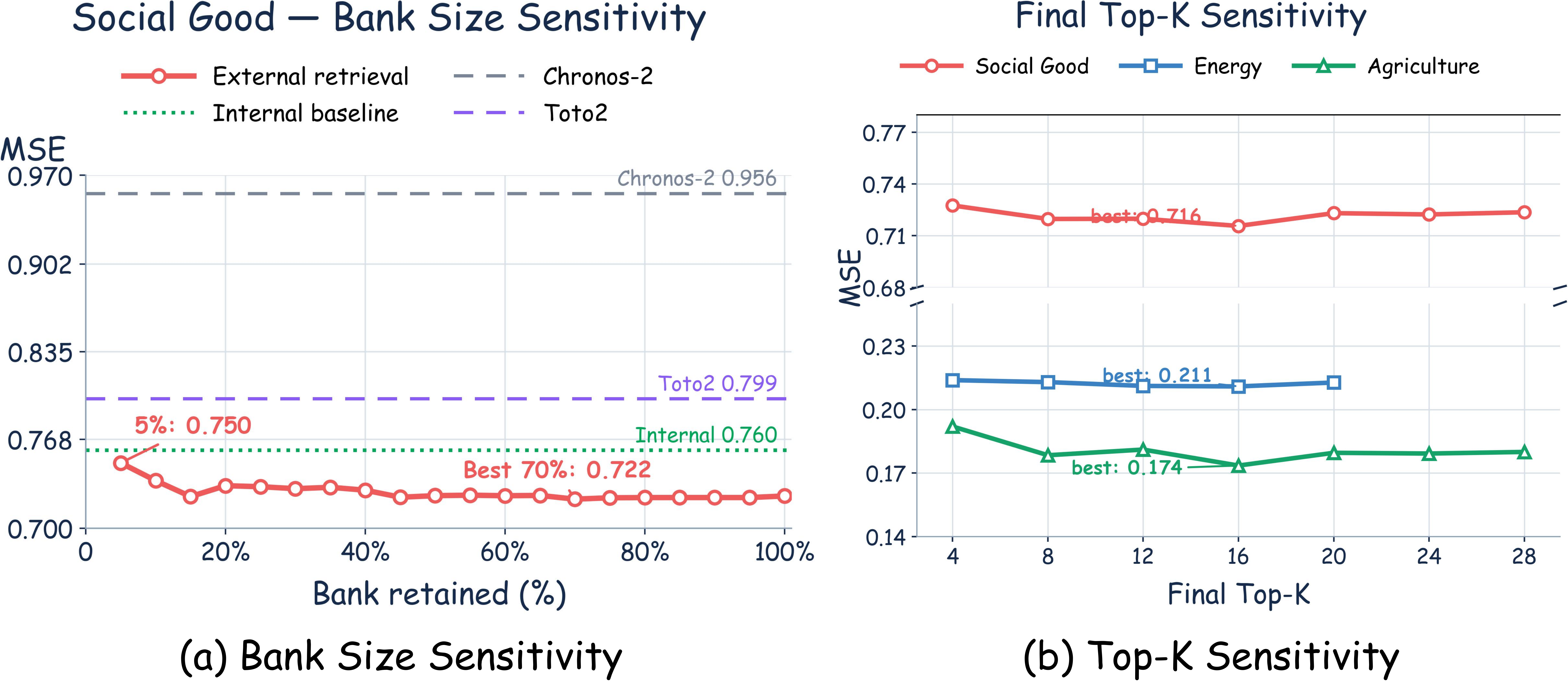}
  \caption{Sensitivity analysis of the retrieval module.}
  \label{fig: retrieval_analysis}
  \vspace{-2mm}
\end{figure*}

We study the robustness of our retrieval mechanism by exploring its
sensitivity to the retrieval bank size and final top-K, as shown in
Figure~\ref{fig: retrieval_analysis}.

For the retrieval bank, \name{} already outperforms the internal baseline
and strong foundation models using only 5\% of the available training
samples.
As the bank size increases, forecasting performance quickly stabilizes
with only minor fluctuations, suggesting that an excessively large
retrieval bank is unnecessary.
In practice, retaining around 20\% of the available samples provides a
favorable trade-off between forecasting performance and retrieval
efficiency.
We further fix the bank size and vary the final top-$k$ from 4 to 24.
The forecasting errors remain within a narrow range across different
values of $k$, indicating that \name{} is insensitive to the exact number
of retrieved samples.
These results demonstrate strong robustness to both retrieval bank size
and top-$k$, substantially reducing the need for careful hyperparameter
tuning.

\section{More Results}
\label{app: More Results}
\subsection{GIFT-Eval Rankings}
\label{app: GIFT-Eval Rankings}
\begin{figure*}[h]
  \centering
  \includegraphics[width=0.55\columnwidth]{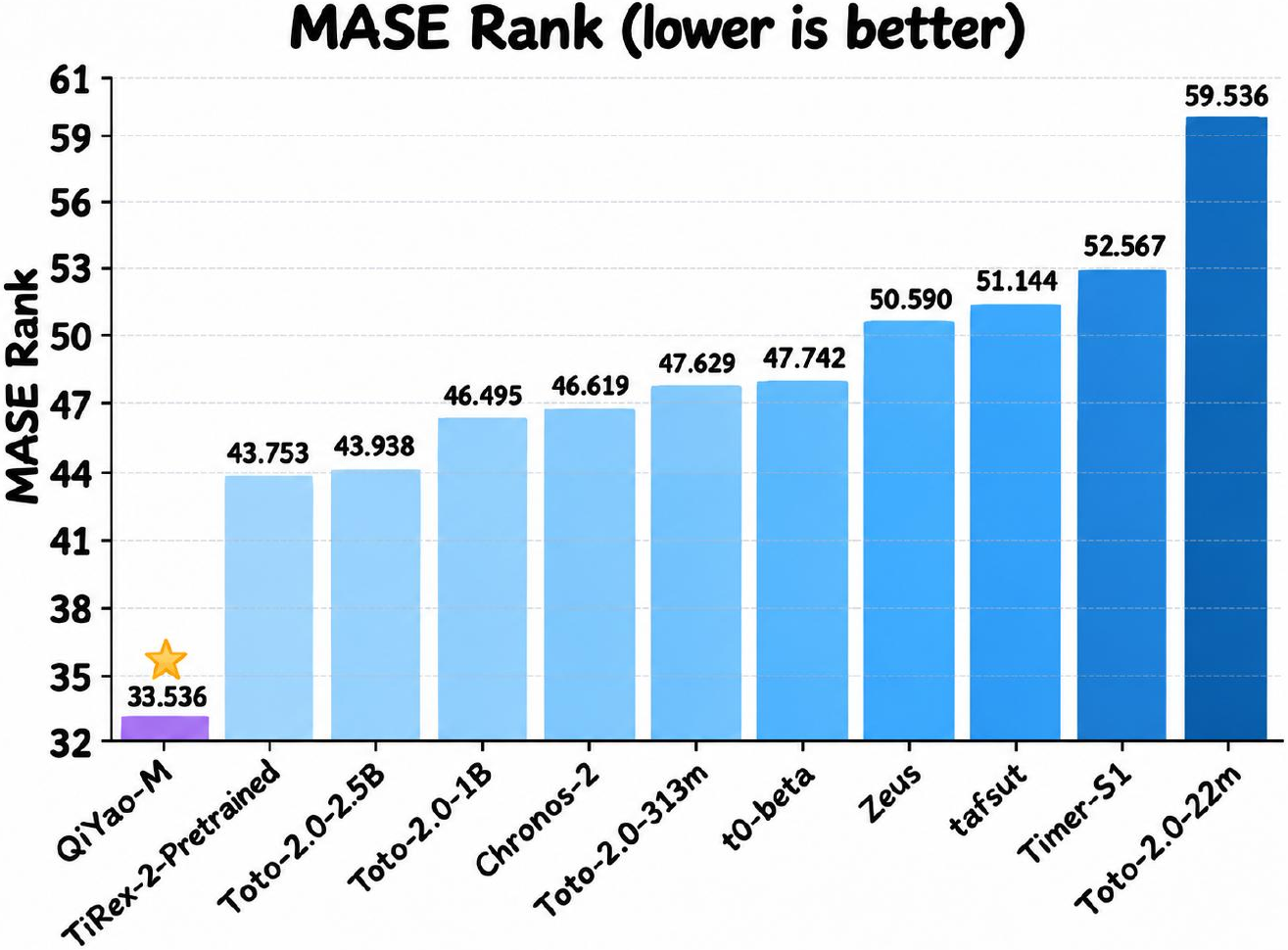}
  \caption{GIFT-Eval MASE Ranking.}
  \label{fig:maserank}
  \vspace{-2mm}
\end{figure*}

\begin{figure*}[h]
  \centering
  \includegraphics[width=0.55\columnwidth]{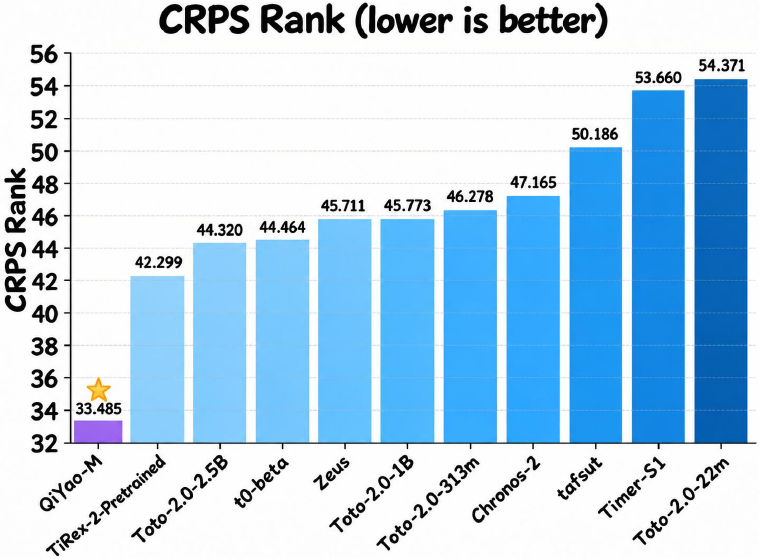}
  \caption{GIFT-Eval CRPS Ranking.}
  \label{fig:crpsrank}
  \vspace{-2mm}
\end{figure*}

To provide a more comprehensive comparison on GIFT-Eval, we further report the ranking results in terms of MASE and CRPS. As shown in Figures~\ref{fig:maserank} and~\ref{fig:crpsrank}, our model consistently achieves favorable rankings across both metrics, demonstrating clear improvements over the baseline forecasting models. These results further validate the effectiveness and robustness of our model under diverse time-series forecasting scenarios.

\subsection{Full Results}
\label{app: Full Results}
We further provide the complete step-wise forecasting results on multimodal datasets. Specifically, we report the performance of different methods at each forecasting horizon, together with their average performance, enabling a more fine-grained comparison across short- and long-term forecasting settings. The detailed results further demonstrate the consistent effectiveness of our model across different prediction horizons and multimodal forecasting benchmarks.


\begin{table*}[!htbp]
\centering
\caption{
Step-wise forecasting results on \textbf{Agriculture}.
Lower MSE and MAE indicate better performance.
The best and second-best results are highlighted in
\rkw{purple} and \rjw{blue}, respectively.
}
\label{tab:agriculture_stepwise}

\renewcommand{\arraystretch}{0.95}
\setlength{\tabcolsep}{8pt}

\resizebox{\textwidth}{!}{%
\begin{tabular}{l|cc|cc|cc|cc|cc}

\toprule
\rowcg
\textbf{Methods}
& \multicolumn{2}{c|}{\textbf{H=6}}
& \multicolumn{2}{c|}{\textbf{H=8}}
& \multicolumn{2}{c|}{\textbf{H=10}}
& \multicolumn{2}{c|}{\textbf{H=12}}
& \multicolumn{2}{c}{\textbf{Avg.}}
\\

& MSE & MAE
& MSE & MAE
& MSE & MAE
& MSE & MAE
& MSE & MAE
\\

\midrule

\rowcg
\multicolumn{11}{c}{
\textit{\textbf{Multimodal End-to-End Models}}
}
\\[-1pt]

\midrule

\methodlogo{Figures/logs/MTS.png}{GPT4MTS}
& 0.161 & 0.257
& 0.207 & 0.288
& 0.230 & \second{0.305}
& 0.301 & 0.342
& 0.225 & 0.298
\\

\methodlogo{Figures/logs/qinghua.png}{CALF}
& 0.142 & 0.250
& 0.195 & 0.285
& 0.350 & 0.370
& 0.314 & 0.355
& 0.250 & 0.315
\\

\methodlogo{Figures/logs/VLM.png}{Time-VLM}
& 0.143 & 0.245
& 0.215 & 0.287
& 0.271 & 0.320
& 0.322 & 0.359
& 0.237 & 0.302
\\

\methodlogo{Figures/logs/TATS.png}{TATS}
& 0.140 & 0.251
& 0.187 & 0.282
& 0.244 & 0.320
& 0.290 & 0.350
& 0.215 & 0.301
\\

\midrule

\rowcg
\multicolumn{11}{c}{
\textit{\textbf{Unimodal Foundation Models}}
}
\\[-1pt]

\midrule

\methodlogo{Figures/logs/Zeus.png}{Zeus}
& 0.122 & 0.237
& 0.173 & 0.279
& 0.228 & 0.315
& 0.289 & 0.348
& 0.203 & 0.295
\\

\methodlogo{Figures/logs/Chronos2.png}{Chronos-2}
& 0.147 & 0.258
& 0.208 & 0.303
& 0.267 & 0.340
& 0.330 & 0.376
& 0.238 & 0.319
\\

\methodlogo{Figures/logs/datadog.png}{Toto-2}
& 0.144 & 0.252
& 0.203 & 0.296
& 0.264 & 0.330
& 0.331 & 0.366
& 0.236 & 0.311
\\

\methodlogo{Figures/logs/PatchTST.png}{PatchTST-r2}
& 0.167 & 0.302
& 0.216 & 0.341
& 0.255 & 0.371
& 0.302 & 0.401
& 0.235 & 0.354
\\

\methodlogo{Figures/logs/TiRex2.png}{TiRex2}
& 0.148 & 0.254
& 0.216 & 0.300
& 0.288 & 0.338
& 0.369 & 0.380
& 0.255 & 0.318
\\

\midrule

\rowcg
\multicolumn{11}{c}{
\textit{\textbf{Multimodal Foundation Models}}
}
\\[-1pt]

\midrule

\methodlogo{Figures/logs/beiyou.png}{ChatTime}
& 0.129 & 0.245
& \second{0.170} & 0.278
& \second{0.215} & 0.306
& \second{0.271} & 0.343
& \second{0.196} & 0.293
\\

\methodlogo{Figures/logs/ECNU.png}{Aurora}
& 0.184 & 0.295
& 0.242 & 0.335
& 0.297 & 0.365
& 0.365 & 0.398
& 0.272 & 0.348
\\

\midrule

\textbf{Ours} (w/o Exo.)
& \second{0.121} & \second{0.233}
& 0.173 & \second{0.275}
& 0.224 & 0.307
& 0.281 & \second{0.340}
& 0.199 & \second{0.288}
\\

\textbf{Ours} (w Exo.)
& \best{0.108} & \best{0.232}
& \best{0.150} & \best{0.268}
& \best{0.192} & \best{0.300}
& \best{0.244} & \best{0.336}
& \best{0.174} & \best{0.284}
\\

\bottomrule
\end{tabular}%
}
\end{table*}


\begin{table*}[!htbp]
\centering
\caption{
Step-wise forecasting results on \textbf{Climate}.
Lower MSE and MAE indicate better performance.
The best and second-best results are highlighted in
\rkw{purple} and \rjw{blue}, respectively.
}
\label{tab:climate_stepwise}

\renewcommand{\arraystretch}{0.95}
\setlength{\tabcolsep}{8pt}

\resizebox{\textwidth}{!}{%
\begin{tabular}{l|cc|cc|cc|cc|cc}

\toprule
\rowcg
\textbf{Horizons}
& \multicolumn{2}{c|}{\textbf{H=6}}
& \multicolumn{2}{c|}{\textbf{H=8}}
& \multicolumn{2}{c|}{\textbf{H=10}}
& \multicolumn{2}{c|}{\textbf{H=12}}
& \multicolumn{2}{c}{\textbf{Avg.}}
\\
\textbf{Metrics}
& MSE & MAE
& MSE & MAE
& MSE & MAE
& MSE & MAE
& MSE & MAE
\\

\midrule

\rowcg
\multicolumn{11}{c}{
\textit{\textbf{Multimodal End-to-End Models}}
}
\\[-1pt]

\midrule

\methodlogo{Figures/logs/MTS.png}{GPT4MTS}
& 1.199 & 0.895
& 1.205 & 0.899
& 1.173 & 0.885
& 1.152 & 0.876
& 1.182 & 0.889
\\

\methodlogo{Figures/logs/qinghua.png}{CALF}
& 1.231 & 0.910
& 1.227 & 0.911
& 1.508 & 0.989
& 1.177 & 0.883
& 1.286 & 0.922
\\

\methodlogo{Figures/logs/VLM.png}{Time-VLM}
& 1.218 & 0.907
& 1.181 & 0.914
& 1.179 & 0.880
& 1.203 & 0.896
& 1.195 & 0.899
\\

\methodlogo{Figures/logs/TATS.png}{TATS}
& 1.194 & 0.897
& 1.178 & 0.886
& 1.170 & 0.881
& 1.179 & 0.885
& 1.180 & 0.887
\\

\midrule

\rowcg
\multicolumn{11}{c}{
\textit{\textbf{Unimodal Foundation Models}}
}
\\[-1pt]

\midrule

\methodlogo{Figures/logs/Zeus.png}{Zeus}
& 0.858 & 0.741
& 0.854 & 0.738
& 0.847 & 0.736
& \second{0.846} & 0.737
& 0.852 & 0.738
\\

\methodlogo{Figures/logs/Chronos2.png}{Chronos-2}
& 0.858 & 0.730
& 0.857 & 0.728
& 0.859 & 0.729
& 0.863 & 0.732
& 0.859 & 0.730
\\

\methodlogo{Figures/logs/datadog.png}{Toto-2}
& 0.844 & 0.725
& 0.850 & 0.728
& 0.858 & 0.731
& 0.861 & 0.733
& 0.853 & 0.729
\\

\methodlogo{Figures/logs/PatchTST.png}{PatchTST-r2}
& 0.842 & \second{0.724}
& \second{0.841} & 0.723
& 0.846 & \second{0.724}
& 0.850 & \best{0.726}
& 0.845 & \second{0.724}
\\

\methodlogo{Figures/logs/TiRex2.png}{TiRex2}
& 0.839 & \best{0.723}
& \second{0.841} & 0.722
& 0.852 & 0.725
& 0.856 & \second{0.727}
& 0.847 & \second{0.724}
\\

\midrule

\rowcg
\multicolumn{11}{c}{
\textit{\textbf{Multimodal Foundation Models}}
}
\\[-1pt]

\midrule

\methodlogo{Figures/logs/beiyou.png}{ChatTime}
& 1.100 & 0.842
& 1.177 & 0.863
& 1.158 & 0.866
& 1.143 & 0.854
& 1.144 & 0.856
\\

\methodlogo{Figures/logs/ECNU.png}{Aurora}
& 0.859 & 0.747
& 0.858 & 0.746
& 0.868 & 0.748
& 0.875 & 0.753
& 0.865 & 0.749
\\

\midrule

\textbf{Ours} (w/o Exo.)
& \second{0.828} & \second{0.724}
& \best{0.822} & \second{0.721}
& \second{0.830} & 0.725
& \best{0.832} & 0.728
& \second{0.828} & 0.725
\\

\textbf{Ours} (w Exo.)
& \best{0.826} & \best{0.723}
& \best{0.822} & \best{0.719}
& \best{0.829} & \best{0.722}
& \best{0.832} & \second{0.727}
& \best{0.827} & \best{0.723}
\\

\bottomrule
\end{tabular}%
}
\end{table*}


\begin{table*}[!htbp]
\centering
\caption{
Step-wise forecasting results on \textbf{Energy}.
Lower MSE and MAE indicate better performance.
The best and second-best results are highlighted in
\rkw{purple} and \rjw{blue}, respectively.
}
\label{tab:energy_stepwise}

\renewcommand{\arraystretch}{0.95}
\setlength{\tabcolsep}{8pt}

\resizebox{\textwidth}{!}{%
\begin{tabular}{l|cc|cc|cc|cc|cc}

\toprule
\rowcg
\textbf{Horizons}
& \multicolumn{2}{c|}{\textbf{H=12}}
& \multicolumn{2}{c|}{\textbf{H=24}}
& \multicolumn{2}{c|}{\textbf{H=36}}
& \multicolumn{2}{c|}{\textbf{H=48}}
& \multicolumn{2}{c}{\textbf{Avg.}}
\\
\textbf{Metrics}
& MSE & MAE
& MSE & MAE
& MSE & MAE
& MSE & MAE
& MSE & MAE
\\

\midrule

\rowcg
\multicolumn{11}{c}{
\textit{\textbf{Multimodal End-to-End Models}}
}
\\[-1pt]

\midrule

\methodlogo{Figures/logs/MTS.png}{GPT4MTS}
& 0.111 & 0.244
& 0.232 & 0.362
& 0.308 & 0.418
& 0.398 & 0.496
& 0.262 & 0.380
\\

\methodlogo{Figures/logs/qinghua.png}{CALF}
& 0.102 & 0.224
& 0.210 & 0.346
& 0.300 & 0.420
& 0.365 & 0.470
& 0.244 & 0.365
\\

\methodlogo{Figures/logs/VLM.png}{Time-VLM}
& 0.114 & 0.253
& 0.227 & 0.359
& 0.309 & 0.410
& 0.390 & 0.475
& 0.260 & 0.374
\\

\methodlogo{Figures/logs/TATS.png}{TATS}
& 0.105 & 0.232
& 0.216 & 0.344
& 0.309 & 0.418
& 0.391 & 0.480
& 0.255 & 0.368
\\

\midrule

\rowcg
\multicolumn{11}{c}{
\textit{\textbf{Unimodal Foundation Models}}
}
\\[-1pt]

\midrule

\methodlogo{Figures/logs/Zeus.png}{Zeus}
& 0.094 & 0.207
& 0.193 & 0.312
& 0.288 & 0.383
& 0.380 & 0.444
& 0.239 & 0.337
\\

\methodlogo{Figures/logs/Chronos2.png}{Chronos-2}
& 0.089 & \second{0.200}
& 0.182 & \best{0.301}
& 0.271 & \best{0.368}
& 0.362 & \second{0.434}
& 0.226 & \second{0.326}
\\

\methodlogo{Figures/logs/datadog.png}{Toto-2}
& 0.091 & 0.203
& 0.190 & 0.315
& 0.275 & 0.383
& 0.367 & 0.445
& 0.231 & 0.337
\\

\methodlogo{Figures/logs/PatchTST.png}{PatchTST-r2}
& 0.087 & 0.210
& 0.182 & 0.319
& 0.284 & 0.399
& 0.426 & 0.484
& 0.245 & 0.353
\\

\methodlogo{Figures/logs/TiRex2.png}{TiRex2}
& 0.088 & 0.202
& \second{0.181} & 0.304
& \second{0.266} & \second{0.373}
& \second{0.360} & 0.441
& \second{0.224} & 0.330
\\

\midrule

\rowcg
\multicolumn{11}{c}{
\textit{\textbf{Multimodal Foundation Models}}
}
\\[-1pt]

\midrule

\methodlogo{Figures/logs/beiyou.png}{ChatTime}
& 0.104 & 0.223
& 0.213 & 0.331
& 0.315 & 0.403
& 0.399 & 0.463
& 0.258 & 0.355
\\

\methodlogo{Figures/logs/ECNU.png}{Aurora}
& 0.117 & 0.245
& 0.226 & 0.354
& 0.292 & 0.409
& 0.383 & 0.472
& 0.255 & 0.370
\\

\midrule

\textbf{Ours} (w/o Exo.)
& \second{0.085} & \second{0.200}
& 0.186 & 0.310
& 0.287 & 0.387
& 0.390 & 0.454
& 0.237 & 0.338
\\

\textbf{Ours} (w Exo.)
& \best{0.081} & \best{0.196}
& \best{0.178} & \second{0.302}
& \best{0.262} & 0.374
& \best{0.322} & \best{0.428}
& \best{0.211} & \best{0.325}
\\

\bottomrule
\end{tabular}%
}
\end{table*}


\begin{table*}[!htbp]
\centering
\caption{
Step-wise forecasting results on \textbf{Health}.
Lower MSE and MAE indicate better performance.
The best and second-best results are highlighted in
\rkw{purple} and \rjw{blue}, respectively.
}
\label{tab:health_stepwise}

\renewcommand{\arraystretch}{0.95}
\setlength{\tabcolsep}{8pt}

\resizebox{\textwidth}{!}{%
\begin{tabular}{l|cc|cc|cc|cc|cc}

\toprule
\rowcg
\textbf{Horizons}
& \multicolumn{2}{c|}{\textbf{H=12}}
& \multicolumn{2}{c|}{\textbf{H=24}}
& \multicolumn{2}{c|}{\textbf{H=36}}
& \multicolumn{2}{c|}{\textbf{H=48}}
& \multicolumn{2}{c}{\textbf{Avg.}}
\\
\textbf{Metrics}
& MSE & MAE
& MSE & MAE
& MSE & MAE
& MSE & MAE
& MSE & MAE
\\

\midrule

\rowcg
\multicolumn{11}{c}{
\textit{\textbf{Multimodal End-to-End Models}}
}
\\[-1pt]

\midrule

\methodlogo{Figures/logs/MTS.png}{GPT4MTS}
& 0.985 & 0.658
& 1.513 & 0.802
& 1.601 & 0.846
& 1.757 & 0.889
& 1.464 & 0.799
\\

\methodlogo{Figures/logs/qinghua.png}{CALF}
& 0.964 & 0.609
& 1.451 & 0.749
& 1.713 & 0.851
& 1.836 & 0.889
& 1.491 & 0.775
\\

\methodlogo{Figures/logs/VLM.png}{Time-VLM}
& 1.198 & 0.727
& 1.491 & 0.839
& 1.867 & 0.967
& 1.702 & 0.907
& 1.565 & 0.860
\\

\methodlogo{Figures/logs/TATS.png}{TATS}
& 0.899 & 0.612
& 1.307 & 0.759
& 1.523 & 0.827
& 1.693 & 0.872
& 1.356 & 0.767
\\

\midrule

\rowcg
\multicolumn{11}{c}{
\textit{\textbf{Unimodal Foundation Models}}
}
\\[-1pt]

\midrule

\methodlogo{Figures/logs/Zeus.png}{Zeus}
& 1.230 & 0.668
& 1.610 & 0.847
& 1.728 & 0.917
& 1.791 & 0.949
& 1.590 & 0.845
\\

\methodlogo{Figures/logs/Chronos2.png}{Chronos-2}
& 0.650 & 0.513
& 0.954 & 0.632
& 1.209 & 0.723
& 1.402 & 0.784
& 1.054 & 0.663
\\

\methodlogo{Figures/logs/datadog.png}{Toto-2}
& 0.702 & 0.501
& 1.038 & 0.637
& 1.260 & 0.723
& 1.437 & 0.787
& 1.109 & 0.662
\\

\methodlogo{Figures/logs/PatchTST.png}{PatchTST-r2}
& 0.643 & 0.512
& 0.960 & 0.636
& 1.119 & 0.697
& 1.233 & 0.743
& 0.989 & 0.647
\\

\methodlogo{Figures/logs/TiRex2.png}{TiRex2}
& 0.754 & 0.518
& 1.116 & 0.635
& 1.236 & 0.685
& 1.280 & \second{0.712}
& 1.097 & 0.638
\\

\midrule

\rowcg
\multicolumn{11}{c}{
\textit{\textbf{Multimodal Foundation Models}}
}
\\[-1pt]

\midrule

\methodlogo{Figures/logs/beiyou.png}{ChatTime}
& 1.266 & 0.716
& 2.087 & 0.952
& 2.622 & 1.097
& 3.137 & 1.240
& 2.278 & 1.001
\\

\methodlogo{Figures/logs/ECNU.png}{Aurora}
& 1.093 & 0.668
& 1.572 & 0.849
& 1.688 & 0.920
& 1.857 & 0.963
& 1.553 & 0.850
\\

\midrule

\textbf{Ours} (w/o Exo.)
& \second{0.585} & \second{0.469}
& \second{0.870} & \second{0.602}
& \second{1.034} & \second{0.664}
& \second{1.170} & \second{0.712}
& \second{0.915} & \second{0.612}
\\

\textbf{Ours} (w Exo.)
& \best{0.574} & \best{0.468}
& \best{0.851} & \best{0.598}
& \best{1.021} & \best{0.659}
& \best{1.156} & \best{0.706}
& \best{0.901} & \best{0.608}
\\

\bottomrule
\end{tabular}%
}
\end{table*}


\begin{table*}[!htbp]
\centering
\caption{
Step-wise forecasting results on \textbf{Social Good}.
Lower MSE and MAE indicate better performance.
The best and second-best results are highlighted in
\rkw{purple} and \rjw{blue}, respectively.
}
\label{tab:social_good_stepwise}

\renewcommand{\arraystretch}{0.95}
\setlength{\tabcolsep}{7pt}

\resizebox{\textwidth}{!}{%
\begin{tabular}{l|cc|cc|cc|cc|cc}

\toprule
\rowcg
\textbf{Horizons}
& \multicolumn{2}{c|}{\textbf{H=6}}
& \multicolumn{2}{c|}{\textbf{H=8}}
& \multicolumn{2}{c|}{\textbf{H=10}}
& \multicolumn{2}{c|}{\textbf{H=12}}
& \multicolumn{2}{c}{\textbf{Avg.}}
\\
\textbf{Metrics}
& MSE & MAE
& MSE & MAE
& MSE & MAE
& MSE & MAE
& MSE & MAE
\\

\midrule

\rowcg
\multicolumn{11}{c}{
\textit{\textbf{Multimodal End-to-End Models}}
}
\\[-1pt]

\midrule

\methodlogo{Figures/logs/MTS.png}{GPT4MTS}
& 0.718 & 0.378
& 0.942 & 0.505
& 0.929 & 0.446
& 1.093 & 0.470
& 0.920 & 0.450
\\

\methodlogo{Figures/logs/qinghua.png}{CALF}
& 0.782 & 0.360
& 0.874 & 0.386
& 0.976 & 0.420
& 0.991 & 0.439
& 0.906 & 0.401
\\

\methodlogo{Figures/logs/VLM.png}{Time-VLM}
& 0.732 & 0.379
& 0.822 & 0.427
& 0.916 & 0.465
& 1.005 & 0.505
& 0.868 & 0.444
\\

\methodlogo{Figures/logs/TATS.png}{TATS}
& 0.753 & 0.370
& 0.875 & 0.409
& 0.991 & 0.459
& 1.053 & 0.474
& 0.918 & 0.428
\\

\midrule

\rowcg
\multicolumn{11}{c}{
\textit{\textbf{Unimodal Foundation Models}}
}
\\[-1pt]

\midrule

\methodlogo{Figures/logs/Zeus.png}{Zeus}
& 0.820 & 0.363
& 0.926 & 0.410
& 1.058 & 0.461
& 1.193 & 0.501
& 0.999 & 0.434
\\

\methodlogo{Figures/logs/Chronos2.png}{Chronos-2}
& 0.734 & 0.312
& 0.855 & 0.362
& 0.959 & 0.408
& 1.060 & 0.453
& 0.902 & 0.384
\\

\methodlogo{Figures/logs/datadog.png}{Toto-2}
& 0.683 & 0.261
& 0.773 & 0.296
& 0.841 & 0.326
& 0.900 & 0.355
& 0.799 & 0.309
\\

\methodlogo{Figures/logs/PatchTST.png}{PatchTST-r2}
& 0.747 & 0.319
& 0.816 & 0.354
& 0.865 & 0.382
& 0.907 & 0.408
& 0.834 & 0.366
\\

\methodlogo{Figures/logs/TiRex2.png}{TiRex2}
& \second{0.650} & 0.277
& \second{0.727} & 0.312
& \second{0.785} & 0.342
& \second{0.832} & 0.369
& \second{0.749} & 0.325
\\

\midrule

\rowcg
\multicolumn{11}{c}{
\textit{\textbf{Multimodal Foundation Models}}
}
\\[-1pt]

\midrule

\methodlogo{Figures/logs/beiyou.png}{ChatTime}
& 1.200 & 0.488
& 1.244 & 0.506
& 1.281 & 0.547
& 1.552 & 0.621
& 1.319 & 0.541
\\

\methodlogo{Figures/logs/ECNU.png}{Aurora}
& 0.701 & 0.442
& 0.804 & 0.493
& 0.886 & 0.543
& 0.960 & 0.587
& 0.838 & 0.516
\\

\midrule

\textbf{Ours} (w/o Exo.)
& 0.675 & \second{0.257}
& 0.739 & \second{0.284}
& 0.789 & \second{0.309}
& 0.835 & \second{0.332}
& 0.760 & \second{0.296}
\\

\textbf{Ours} (w Exo.)
& \best{0.634} & \best{0.253}
& \best{0.702} & \best{0.283}
& \best{0.753} & \best{0.305}
& \best{0.800} & \best{0.331}
& \best{0.722} & \best{0.293}
\\

\bottomrule
\end{tabular}%
}
\end{table*}

\begin{table*}[h]
\centering
\caption{
Step-wise forecasting results on \textbf{TAOBAO-fashion}.
Lower MSE and MAE indicate better performance.
The best and second-best results are highlighted in
\rkw{purple} and \rjw{blue}, respectively.
}
\label{tab:taobao_stepwise}

\renewcommand{\arraystretch}{0.95}
\setlength{\tabcolsep}{5pt}

\resizebox{\textwidth}{!}{%
\begin{tabular}{l|cc|cc|cc|cc|cc|cc}

\toprule

\rowcg
\textbf{Horizons}
& \multicolumn{2}{c|}{\textbf{H=1}}
& \multicolumn{2}{c|}{\textbf{H=7}}
& \multicolumn{2}{c|}{\textbf{H=14}}
& \multicolumn{2}{c|}{\textbf{H=21}}
& \multicolumn{2}{c|}{\textbf{H=28}}
& \multicolumn{2}{c}{\textbf{Avg.}}
\\
\textbf{Metrics}
& MSE & MAE
& MSE & MAE
& MSE & MAE
& MSE & MAE
& MSE & MAE
& MSE & MAE
\\

\midrule


\rowcg
\multicolumn{13}{c}{
\textit{\textbf{Multimodal End-to-End Models}}
}
\\[-1pt]

\midrule

\methodlogo{Figures/logs/MTS.png}{GPT4MTS}
& 0.384 & 0.189
& 0.473 & 0.232
& 0.613 & 0.297
& 0.640 & 0.293
& 0.559 & 0.272
& 0.534 & 0.257
\\

\methodlogo{Figures/logs/qinghua.png}{CALF}
& 0.420 & 0.230
& 0.488 & 0.240
& 0.544 & 0.263
& 0.536 & 0.254
& 0.570 & 0.263
& 0.512 & 0.250
\\

\methodlogo{Figures/logs/VLM.png}{Time-VLM}
& 0.419 & 0.226
& 0.489 & 0.257
& 0.555 & 0.290
& 0.568 & 0.290
& 0.609 & 0.289
& 0.528 & 0.271
\\

\methodlogo{Figures/logs/TATS.png}{TATS}
& 0.443 & 0.242
& 0.515 & 0.258
& 0.538 & 0.273
& 0.548 & 0.267
& 0.579 & 0.274
& 0.525 & 0.263
\\

\midrule


\rowcg
\multicolumn{13}{c}{
\textit{\textbf{Unimodal Foundation Models}}
}
\\[-1pt]

\midrule

\methodlogo{Figures/logs/Zeus.png}{Zeus}
& 0.376 & 0.153
& 0.394 & \second{0.161}
& 0.426 & \best{0.172}
& \second{0.450} & \best{0.181}
& 0.477 & \best{0.191}
& 0.425 & \second{0.172}
\\

\methodlogo{Figures/logs/Chronos2.png}{Chronos-2}
& 0.378 & 0.153
& 0.411 & 0.169
& 0.449 & 0.186
& 0.477 & 0.199
& 0.509 & 0.212
& 0.445 & 0.184
\\

\methodlogo{Figures/logs/datadog.png}{Toto-2}
& 0.371 & 0.151
& 0.406 & 0.166
& 0.443 & 0.182
& 0.471 & 0.197
& 0.505 & 0.213
& 0.439 & 0.182
\\

\methodlogo{Figures/logs/PatchTST.png}{PatchTST-r2}
& 0.371 & 0.163
& 0.503 & 0.186
& 0.551 & 0.201
& 0.585 & 0.216
& 0.621 & 0.232
& 0.526 & 0.200
\\

\methodlogo{Figures/logs/TiRex2.png}{TiRex2}
& \second{0.358} & \second{0.150}
& 0.400 & \best{0.160}
& 0.434 & \best{0.172}
& 0.459 & \second{0.182}
& 0.487 & \second{0.193}
& \second{0.427} & \best{0.171}
\\
\midrule

\rowcg
\multicolumn{13}{c}{
\textit{\textbf{Multimodal Foundation Models}}
}
\\[-1pt]

\midrule

\methodlogo{Figures/logs/beiyou.png}{ChatTime}
& 0.544 & 0.167
& 0.432 & 0.185
& 0.357 & 0.200
& 0.527 & 0.219
& 0.537 & 0.231
& 0.479 & 0.201
\\

\methodlogo{Figures/logs/ECNU.png}{Aurora}
& 0.365 & 0.169
& 0.408 & 0.192
& 0.448 & 0.214
& 0.475 & 0.229
& 0.512 & 0.247
& 0.442 & 0.210
\\

\midrule


\textbf{Ours} (w/o Exo.)
& \best{0.351} & \best{0.149}
& \second{0.391} & \second{0.161}
& 0.424 & \best{0.172}
& \best{0.447} & \second{0.182}
& \second{0.473} & \second{0.193}
& \best{0.417} & \best{0.171}
\\

\textbf{Ours} (w Exo.)
& 0.469 & 0.152
& \best{0.359} & 0.160
& \second{0.324} & \second{0.173}
& 0.460 & 0.185
& \best{0.472} & 0.200
& \best{0.417} & 0.174
\\

\bottomrule

\end{tabular}%
}

\end{table*}


\begin{table*}[h]
\centering
\caption{
Step-wise forecasting results on \textbf{Tianchi}.
Lower MSE and MAE indicate better performance.
The best and second-best results are highlighted in
\rkw{purple} and \rjw{blue}, respectively.
}
\label{tab:tianchi_stepwise}

\renewcommand{\arraystretch}{0.95}
\setlength{\tabcolsep}{3.5pt}

\resizebox{\textwidth}{!}{%
\begin{tabular}{l|cc|cc|cc|cc|cc|cc}

\toprule

\rowcg
\textbf{Horizons}
& \multicolumn{2}{c|}{\textbf{H=1}}
& \multicolumn{2}{c|}{\textbf{H=7}}
& \multicolumn{2}{c|}{\textbf{H=14}}
& \multicolumn{2}{c|}{\textbf{H=21}}
& \multicolumn{2}{c|}{\textbf{H=28}}
& \multicolumn{2}{c}{\textbf{Avg.}}
\\
\textbf{Metrics}
& MSE & MAE
& MSE & MAE
& MSE & MAE
& MSE & MAE
& MSE & MAE
& MSE & MAE
\\

\midrule


\rowcg
\multicolumn{13}{c}{
\textit{\textbf{Multimodal End-to-End Models}}
}
\\[-1pt]

\midrule

\methodlogo{Figures/logs/MTS.png}{GPT4MTS}
& \best{1.931} & 0.156
& 1.384 & 0.138
& 1.112 & 0.154
& 1.013 & 0.141
& 1.162 & 0.145
& 1.320 & 0.147
\\

\methodlogo{Figures/logs/qinghua.png}{CALF}
& 2.017 & 0.128
& 1.346 & 0.108
& 1.065 & 0.111
& 0.996 & 0.114
& 1.176 & 0.134
& 1.320 & 0.119
\\

\methodlogo{Figures/logs/VLM.png}{Time-VLM}
& \second{1.982} & 0.124
& 1.416 & 0.135
& 1.217 & 0.155
& 1.027 & 0.129
& 1.200 & 0.140
& 1.368 & 0.137
\\

\methodlogo{Figures/logs/TATS.png}{TATS}
& 2.035 & 0.135
& 1.411 & 0.134
& 1.165 & 0.138
& 1.052 & 0.133
& 1.206 & 0.139
& 1.374 & 0.136
\\

\midrule


\rowcg
\multicolumn{13}{c}{
\textit{\textbf{Unimodal Foundation Models}}
}
\\[-1pt]

\midrule

\methodlogo{Figures/logs/Zeus.png}{Zeus}
& 2.143 & 0.085
& 1.339 & 0.074
& 1.188 & 0.083
& 1.065 & 0.083
& 1.154 & 0.091
& 1.378 & 0.083
\\

\methodlogo{Figures/logs/Chronos2.png}{Chronos-2}
& 2.707 & 0.089
& 1.337 & 0.075
& 1.129 & 0.082
& 0.975 & 0.082
& 1.119 & 0.089
& 1.453 & 0.083
\\

\methodlogo{Figures/logs/datadog.png}{Toto-2}
& 2.545 & 0.096
& 2.882 & 0.107
& 119.296 & 0.303
& 54.950 & 0.224
& 28.415 & 0.194
& 41.618 & 0.185
\\

\methodlogo{Figures/logs/PatchTST.png}{PatchTST-r2}
& 2.150 & 0.087
& 1.337 & 0.083
& 1.396 & 0.097
& 0.974 & 0.091
& 1.197 & 0.099
& 1.411 & 0.091
\\

\methodlogo{Figures/logs/TiRex2.png}{TiRex2}
& 2.122 & \best{0.078}
& \second{1.324} & \best{0.069}
& \best{1.033} & \best{0.073}
& \best{0.936} & \best{0.072}
& \best{1.079} & \best{0.079}
& \best{1.299} & \best{0.074}
\\

\midrule


\rowcg
\multicolumn{13}{c}{
\textit{\textbf{Multimodal Foundation Models}}
}
\\[-1pt]

\midrule

\methodlogo{Figures/logs/beiyou.png}{ChatTime}
& 2.478 & 0.090
& 1.372 & 0.078
& 2.045 & 0.094
& 1.306 & 0.087
& 1.179 & 0.088
& 1.676 & 0.087
\\

\methodlogo{Figures/logs/ECNU.png}{Aurora}
& 2.219 & 0.086
& 1.321 & 0.082
& 1.204 & 0.088
& 0.998 & 0.092
& 1.272 & 0.102
& 1.403 & 0.090
\\

\midrule


\textbf{Ours} (w/o Exo.)
& 2.563 & 0.089
& 1.328 & 0.072
& 1.053 & 0.078
& 1.052 & 0.081
& 1.121 & 0.086
& 1.423 & 0.081
\\

\textbf{Ours} (w Exo.)
& 2.132 & \second{0.081}
& \best{1.310} & \second{0.070}
& \best{1.021} & \second{0.074}
& \second{0.963} & \second{0.075}
& \second{1.084} & \second{0.081}
& \second{1.302} & \second{0.076}
\\

\bottomrule

\end{tabular}%
}

\end{table*}


\begin{table*}[h]
\centering
\caption{
Step-wise forecasting results on \textbf{HS300}.
Lower MSE and MAE indicate better performance.
The best and second-best results are highlighted in
\rkw{purple} and \rjw{blue}, respectively.
}
\label{tab:hs300_stepwise}

\renewcommand{\arraystretch}{0.95}
\setlength{\tabcolsep}{10pt}

\resizebox{\textwidth}{!}{%
\begin{tabular}{l|cc|cc|cc|cc}

\toprule

\rowcg
\textbf{Horizons}
& \multicolumn{2}{c|}{\textbf{H=24}}
& \multicolumn{2}{c|}{\textbf{H=48}}
& \multicolumn{2}{c|}{\textbf{H=96}}
& \multicolumn{2}{c}{\textbf{Avg.}}
\\
\textbf{Metrics}
& MSE & MAE
& MSE & MAE
& MSE & MAE
& MSE & MAE
\\

\midrule


\rowcg
\multicolumn{9}{c}{
\textit{\textbf{Multimodal End-to-End Models}}
}
\\[-1pt]

\midrule

\methodlogo{Figures/logs/MTS.png}{GPT4MTS}
& 0.321 & 0.352
& 0.626 & 0.493
& 1.392 & 0.759
& 0.780 & 0.535
\\

\methodlogo{Figures/logs/qinghua.png}{CALF}
& 0.283 & 0.333
& 0.574 & 0.475
& \best{1.165} & \second{0.696}
& 0.674 & 0.502
\\

\methodlogo{Figures/logs/VLM.png}{Time-VLM}
& 0.287 & 0.347
& \best{0.500} & \best{0.456}
& 1.200 & \best{0.693}
& \best{0.662} & 0.499
\\

\methodlogo{Figures/logs/TATS.png}{TATS}
& 0.281 & 0.330
& 0.561 & \second{0.461}
& 1.309 & 0.737
& 0.717 & 0.509
\\

\midrule


\rowcg
\multicolumn{9}{c}{
\textit{\textbf{Unimodal Foundation Models}}
}
\\[-1pt]

\midrule

\methodlogo{Figures/logs/Zeus.png}{Zeus}
& 0.306 & 0.341
& 0.625 & 0.493
& 1.333 & 0.739
& 0.755 & 0.524
\\

\methodlogo{Figures/logs/Chronos2.png}{Chronos-2}
& 0.397 & 0.329
& 0.728 & 0.475
& 1.306 & 0.699
& 0.810 & 0.501
\\

\methodlogo{Figures/logs/datadog.png}{Toto-2}
& 0.501 & 0.331
& 0.773 & 0.475
& 1.382 & 0.716
& 0.885 & 0.507
\\

\methodlogo{Figures/logs/PatchTST.png}{PatchTST-r2}
& 0.292 & 0.323
& 0.635 & 0.481
& 1.529 & 0.759
& 0.819 & 0.521
\\

\methodlogo{Figures/logs/TiRex2.png}{TiRex2}
& \second{0.277} & 0.322
& 0.565 & 0.473
& \second{1.179} & 0.701
& 0.674 & 0.499
\\

\midrule


\rowcg
\multicolumn{9}{c}{
\textit{\textbf{Multimodal Foundation Models}}
}
\\[-1pt]

\midrule

\methodlogo{Figures/logs/beiyou.png}{ChatTime}
& 0.628 & 0.492
& 1.115 & 0.669
& 1.880 & 0.898
& 1.208 & 0.686
\\

\methodlogo{Figures/logs/ECNU.png}{Aurora}
& 0.332 & 0.359
& 0.625 & 0.496
& 1.251 & 0.716
& 0.736 & 0.524
\\

\midrule


\textbf{Ours} (w/o Exo.)
& \best{0.273} & \second{0.319}
& 0.567 & 0.466
& 1.232 & 0.703
& 0.690 & \second{0.496}
\\

\textbf{Ours} (w Exo.)
& \best{0.273} & \best{0.318}
& \second{0.549} & 0.463
& 1.194 & 0.698
& \second{0.672} & \best{0.493}
\\

\bottomrule

\end{tabular}%
}

\end{table*}


\begin{table*}[h]
\centering
\caption{
Step-wise forecasting results on \textbf{SP500}.
Lower MSE and MAE indicate better performance.
The best and second-best results are highlighted in
\rkw{purple} and \rjw{blue}, respectively.
}
\label{tab:sp500_stepwise}

\renewcommand{\arraystretch}{0.95}
\setlength{\tabcolsep}{10pt}

\resizebox{\textwidth}{!}{%
\begin{tabular}{l|cc|cc|cc|cc}

\toprule

\rowcg
\textbf{Horizons}
& \multicolumn{2}{c|}{\textbf{H=24}}
& \multicolumn{2}{c|}{\textbf{H=48}}
& \multicolumn{2}{c|}{\textbf{H=96}}
& \multicolumn{2}{c}{\textbf{Avg.}}
\\
\textbf{Metrics}
& MSE & MAE
& MSE & MAE
& MSE & MAE
& MSE & MAE
\\

\midrule


\rowcg
\multicolumn{9}{c}{
\textit{\textbf{Multimodal End-to-End Models}}
}
\\[-1pt]

\midrule

\methodlogo{Figures/logs/MTS.png}{GPT4MTS}
& 0.484 & 0.503
& 0.723 & 0.635
& 1.095 & 0.786
& 0.767 & 0.642
\\

\methodlogo{Figures/logs/qinghua.png}{CALF}
& 0.368 & 0.434
& \second{0.598} & 0.564
& 1.262 & 0.853
& 0.743 & 0.617
\\

\methodlogo{Figures/logs/VLM.png}{Time-VLM}
& 0.395 & 0.449
& 0.760 & 0.629
& 1.114 & 0.776
& 0.756 & 0.618
\\

\methodlogo{Figures/logs/TATS.png}{TATS}
& 0.365 & 0.430
& 0.617 & 0.570
& \best{1.014} & \best{0.742}
& \second{0.665} & 0.581
\\

\midrule


\rowcg
\multicolumn{9}{c}{
\textit{\textbf{Unimodal Foundation Models}}
}
\\[-1pt]

\midrule

\methodlogo{Figures/logs/Zeus.png}{Zeus}
& 0.385 & 0.427
& 0.728 & 0.595
& 1.461 & 0.862
& 0.858 & 0.628
\\

\methodlogo{Figures/logs/Chronos2.png}{Chronos-2}
& 0.354 & 0.408
& 0.651 & 0.559
& 1.147 & 0.765
& 0.717 & 0.577
\\

\methodlogo{Figures/logs/datadog.png}{Toto-2}
& \second{0.332} & \best{0.395}
& 0.620 & \second{0.544}
& 1.157 & 0.772
& 0.703 & 0.571
\\

\methodlogo{Figures/logs/PatchTST.png}{PatchTST-r2}
& 0.471 & 0.434
& 0.948 & 0.609
& 1.822 & 0.858
& 1.080 & 0.634
\\

\methodlogo{Figures/logs/TiRex2.png}{TiRex2}
& 0.344 & 0.405
& 0.624 & 0.557
& 1.159 & 0.779
& 0.709 & 0.580
\\

\midrule


\rowcg
\multicolumn{9}{c}{
\textit{\textbf{Multimodal Foundation Models}}
}
\\[-1pt]

\midrule

\methodlogo{Figures/logs/beiyou.png}{ChatTime}
& 5.214 & 1.647
& 5.719 & 1.747
& 7.022 & 1.944
& 5.985 & 1.779
\\

\methodlogo{Figures/logs/ECNU.png}{Aurora}
& 0.449 & 0.476
& 0.761 & 0.622
& 1.376 & 0.851
& 0.862 & 0.649
\\

\midrule


\textbf{Ours} (w/o Exo.)
& \best{0.329} & \second{0.396}
& 0.612 & 0.545
& 1.142 & 0.767
& 0.694 & \second{0.569}
\\

\textbf{Ours} (w Exo.)
& \best{0.329} & 0.398
& \best{0.591} & \best{0.542}
& \second{1.061} & \second{0.744}
& \best{0.660} & \best{0.561}
\\

\bottomrule

\end{tabular}%
}

\end{table*}

\end{document}